\documentclass[11pt,a4paper]{article}
\usepackage[hyperref]{emnlp2020}
\usepackage{times}
\usepackage{latexsym}

\usepackage{todonotes}
\usepackage{graphicx}
\usepackage{caption}
\usepackage[list=false]{subcaption}
\usepackage{comment}
\usepackage{amsfonts}
\usepackage{mathtools, cases}

\DeclareMathOperator*{\argmax}{argmax}
\usepackage{footnote}
\usepackage{hyperref}
\usepackage{tabularx}
\usepackage{multicol}
\usepackage{tablefootnote}
\usepackage{color, colortbl}
\usepackage{pifont}

\definecolor{Gray}{gray}{0.9}

\definecolor{cambridgeblue}{rgb}{0.64, 0.76, 0.68}
\definecolor{blue(ncs)}{rgb}{0.0, 0.53, 0.74}
\definecolor{cadetblue}{rgb}{0.37, 0.62, 0.63}
\definecolor{cadmiumgreen}{rgb}{0.0, 0.42, 0.24}
\definecolor{ao(english)}{rgb}{0.0, 0.5, 0.0}
\definecolor{alizarin}{rgb}{0.82, 0.1, 0.26}
\definecolor{vermilion}{rgb}{0.89, 0.26, 0.2}

\usepackage{tikz}

\usetikzlibrary{shapes.geometric}

\usepackage{forest}

\usepackage{microtype}

\aclfinalcopy 

\title{KeyGen2Vec: Learning Document Embedding via Multi-label Keyword Generation in Question-Answering}

\author{Iftitahu Ni'mah\textsuperscript{\ding{168},\ding{171}} \hspace{1.5mm} Samaneh Khoshrou\textsuperscript{\ding{168}} \hspace{1.5mm} Vlado Menkovski\textsuperscript{\ding{168}} \hspace{1.5mm} Mykola Pechenizkiy\textsuperscript{\ding{168}}    \\
{\normalsize{\textsuperscript{\ding{168}} Eindhoven University of Technology \hspace{1.5mm}\textsuperscript{\ding{171}} BRIN Indonesia}} \\
\texttt{\small{\{i.nimah, v.menkovski, m.pechenizkiy\}@tue.nl}}}

\date{}

\begin{document}
\maketitle
\begin{abstract}


Representing documents into high dimensional embedding space while preserving the structural similarity between document sources has been an ultimate goal for many works on text representation learning. Current embedding models, however, mainly rely on the availability of label supervision to increase the expressiveness of the resulting embeddings. In contrast, unsupervised embeddings are cheap, but they often cannot capture implicit structure in target corpus, particularly for samples that come from different distribution with the pretraining source. 

Our study aims to loosen up the dependency on label supervision by learning document embeddings via Sequence-to-Sequence (Seq2Seq) text generator. Specifically, we reformulate keyphrase generation task into multi-label keyword generation in community-based Question Answering (cQA). Our empirical results show that \textbf{KeyGen2Vec} in general is superior than multi-label keyword classifier by up to 14.7\% based on Purity, Normalized Mutual Information (NMI), and F1-Score metrics. Interestingly, although in general the absolute advantage of learning embeddings through label supervision is highly positive across evaluation datasets, \textbf{KeyGen2Vec} is shown to be competitive with classifier that exploits topic label supervision in Yahoo\! cQA with larger number of latent topic labels. \footnote{The empirical study was completed in 2020 at Eindhoven University of Technology.}



\end{abstract}

\section{Introduction}
\label{sec:intro}
\vspace{-0.5em}

\begin{figure}[!ht]
     \centering
     \begin{subfigure}[ht]{0.235\textwidth}
         \centering
         \includegraphics[width=1.5in,height=1in]{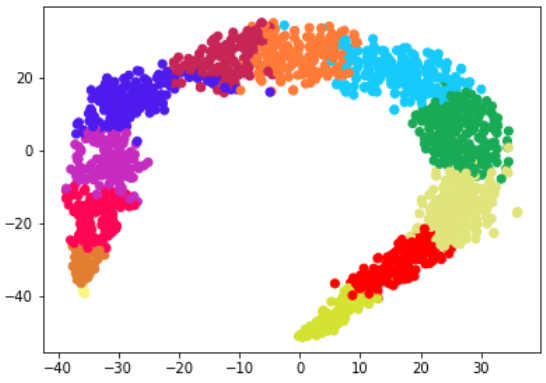}
         \caption{Doc2Vec clusters}
         \label{fig:cluster_vis1}
     \end{subfigure}
     \begin{subfigure}[ht]{0.235\textwidth}
         \centering
         \includegraphics[width=1.5in,height=1in]{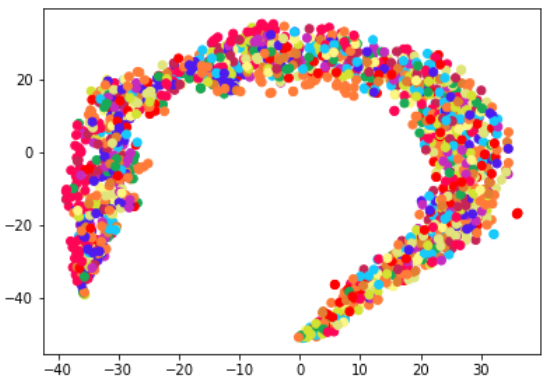}
         \caption{Fitting labels}
         \label{fig:cluster_vis2}
     \end{subfigure}
     \begin{subfigure}[ht]{0.235\textwidth}
         \centering
         \includegraphics[width=1.5in,height=1in]{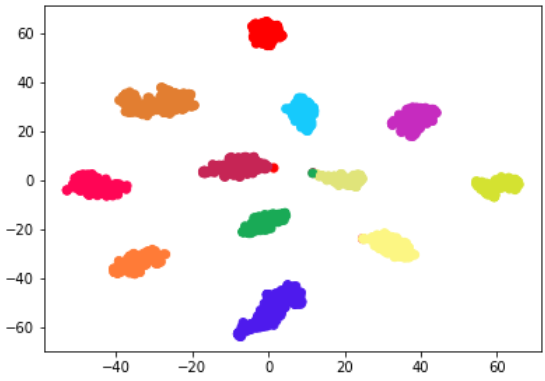}
         \caption{Supervised clusters}
         \label{fig:cluster_vis3}
     \end{subfigure}
     \begin{subfigure}[ht]{0.235\textwidth}
         \centering
         \includegraphics[width=1.5in,height=1in]{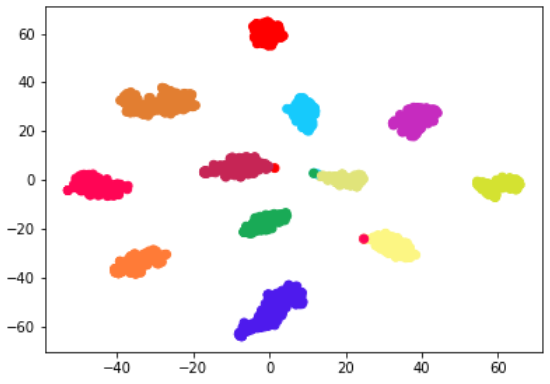}
         \caption{Fitting labels}
         \label{fig:cluster_vis4}
     \end{subfigure}
     \vspace{-.5em}
        \caption{The expressiveness between unsupervised and supervised document embeddings\footnotemark. Upper: \textbf{unsupervised} embeddings via Doc2Vec. Lower: \textbf{supervised} embeddings via a multi-class classifier.}
        \label{fig:cluster_vis}
    \vspace{-1.6em}
\end{figure}

Keywords or \emph{tags} have been widely used in community-driven Question-Answer (cQA) systems and many online platforms, such as Twitter, and online bibliographic databases and search engines, as metadata describing sub-topics of an article. Obtaining keywords as supervisory information for training any machine learning models is considered to be cheaper than obtaining topic labels since there exists automated keyword extraction methods, e.g. TfIdf, TextRank \cite{mihalcea-tarau-2004-textrank}, SingleRank \cite{wan2008singlerank}, Maui \cite{medelyan-etal-2009-human}, which in practice is often accompanied by human validation to control the quality of keyword labels. 

\footnotetext{2-D Projection is based on t-SNE. Points are colored according to the predicted cluster ids $\Omega$ (left) and exact classes $\mathbb{C}$ (right). For the consistency plotting of cluster membership colors, exact classes was mapped to prediction cluster  ids by ``Hungarian'' algorithm.}

\begin{figure*}[!ht]
     \centering
	\includegraphics[width=.9\linewidth]{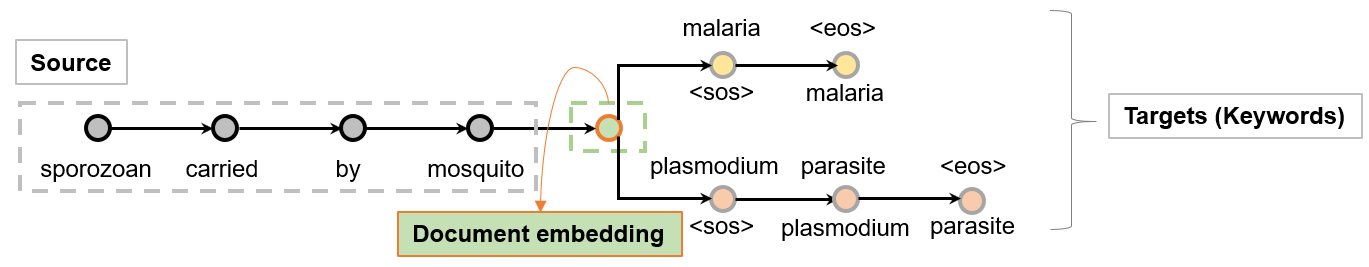}
	\caption{\textbf{KeyGen2Vec}. Model is trained in \emph{autoregressive} mode. In teacher forcing mode, $t-1$ shifted version of keyword is given to the model as input for decoder. Circles represent RNN states in encoder (left/source) and decoder (right/target) network. }
	\label{fig:KeyGen2Vec}
\end{figure*}


Despite many potential benefits of keywords or tags, such as providing predictable patterns \cite{golder2005structure,golder2006usage,nimah2019bsdar,nimah-etal-2021-protoinfomax-prototypical}; knowledge organization and resource discovery \cite{macgregor2006collaborative,ames2007we}; improving retrieval performance \cite{hotho2006information,nimah2019bsdar}, less attention has been paid to incorporate keywords as a condition to embed documents in high dimensional embedding space. Existing approaches that incorporate keywords, tags, or phrases as additional semantic knowledge for document clustering can be divided into two: (1) approaches that focus on improving the quality of document embeddings during training \cite{pu2015topic,sato-etal-2017-distributed}; and (2) approaches that focus on improving the clustering algorithm or subsequent tasks by providing additional post-pipelines \cite{ramage2009clustering,rosa2011topical,dao2018descriptive} to intensify the expressiveness of representations in latent space, such that semantically similar points in that space are close together compared to dissimilar points. However, these works depend on multiple pipelines, which consequently hinder their reproducibility and adaptation as end-to-end system in many real world NLP applications.




\begin{figure*}[!ht]

    \centering
    \tikzset{>=latex}
\forestset{
  stepwise/.style n args=2{
    edge path={
      \noexpand\path [draw, \forestoption{edge}] (!u.parent anchor) |- +(#1,#2) |- (.child anchor)\forestoption{edge label};
    }
  },
  my shading/.style={
    for tree={
      text/.wrap pgfmath arg={black!##1!#1}{10*level()},
      edge/.wrap pgfmath arg={->, draw=black!##1!#1}{10*level()},
    },
  },
}
\resizebox{\textwidth}{!}{
\begin{forest}
  for tree={
    edge=->,
    grow=east,
    align=left,
    child anchor=west,
    edge path={
      \noexpand\path [draw, \forestoption{edge}] (!u.parent anchor) |- (.child anchor)\forestoption{edge label};
    },
    font=\sffamily
  }
  [Approaches
    [3. Distributed
        [e. Discrete predictor (Supervised)
            [2. Multi-Label Classifier; {[} {4, 5} {]} in table \ref{tab:all_models}, name=multilbl]
            [1. Multi-Class Classifier; {[} {3} {]} in table \ref{tab:all_models}, name=upperbound]
        ]
        [d. Pretrained Sentence Encoder: S-BERT \cite{reimers-gurevych-2019-sentence}; {[} {6} {]} in table \ref{tab:all_models}]
        [c. Paragraph Vector (Doc2Vec) \cite{le2014distributed}
        {[} {8,9} {]} in table \ref{tab:all_models}]
        [b. Seq2Seq
            [\Large 2. \textbf{KeyGen2Vec} \\
            {[} {1} {]} in table \ref{tab:all_models}, my shading=vermilion, name=framework]
            [1. Autoencoder \\
            \cite{sutskever2014sequence,cho2014learning}; {[} {2} {]} in table \ref{tab:all_models}]
        ]
        [a. Bottom-Up
            [2. Covariance Matrix \cite{nikolentzos-etal-2017-multivariate,torki-2018-document}
                [b. Neural Word Embedding;\\
                {[} {17,18,20} {]} in table \ref{tab:all_models}
                ]
                [a. PMI; {[} {19} {]} in table \ref{tab:all_models}]
            ]
            [1. Mean Embedding
                [d. Pretrained $\rightarrow$ Wiki2Vec \cite{yamada2020wikipedia2vec}\\ and fastText \cite{bojanowski2017enriching}; {[} {11, 13} {]} in table \ref{tab:all_models}]
                [c. Pretrained $\rightarrow$ GloVe \\ \cite{pennington2014glove}; {[} {10, 12} {]} in table \ref{tab:all_models}]
                [b. Trainable $\rightarrow$  Word2Vec\\ \cite{mikolov2013distributed,mikolov2013efficient}; {[} {14, 15} {]} in table \ref{tab:all_models}
                ]
                [a. Pointwise Mutual Information (PMI)\\
                \cite{levy2014neural}; {[} {16} {]} in table \ref{tab:all_models}
                ]
            ]
        ]
    ]
    [2. Probabilistic
        [Dirichlet-based Topic Model (LDA) \\ \cite{blei2010topic,blei2012probabilistic}; \\
        {[}7{]} in table \ref{tab:all_models}
        ]
    ]
    [1. Non-Distributed
        [TfIdf \cite{salton1975vector}; {[}21{]} in table \ref{tab:all_models}
        ]
    ]
  ]
  \node [draw=vermilion, solid, line width=1mm, inner sep=.4em, fit={(framework)}, label={[align=left]0: \textbf{\LARGE {\color{vermilion}{}Proposed Framework}} }] {};
\end{forest}
}
    \caption{Document embedding approaches in this study.}
    \label{fig:doc_rep_app}
\end{figure*}
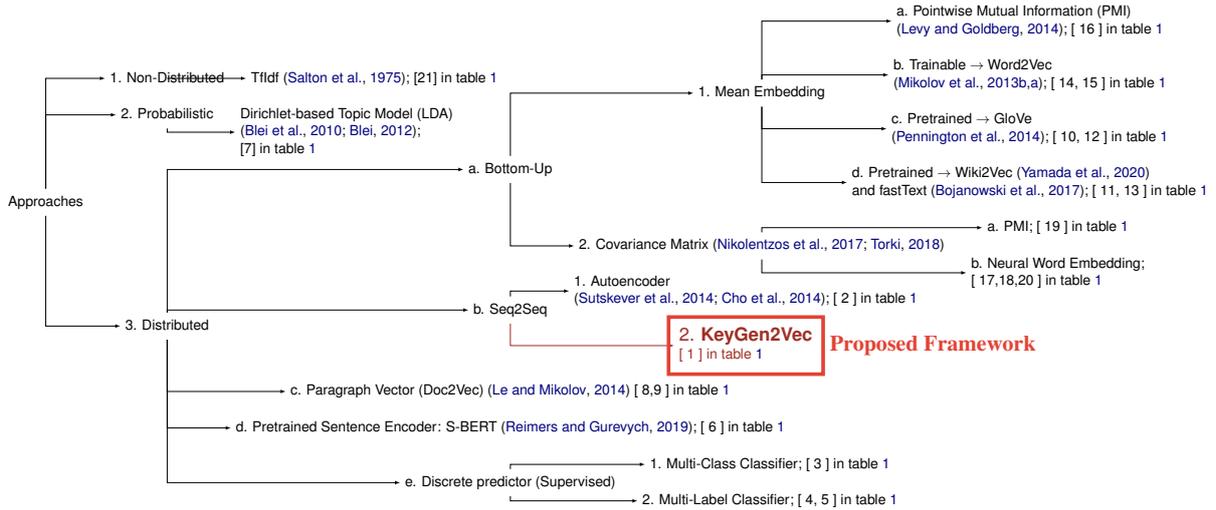

Our work mainly focuses on topical clustering of cQA archives as a subsequent task to evaluate currently available document embedding approaches, including the proposed \textbf{KeyGen2Vec} framework. As a motivating example, Figure~\ref{fig:cluster_vis1}-\ref{fig:cluster_vis2} illustrate how unsupervised-based embeddings are likely random, indicating the model's incapability to capture semantic aspects such as latent topics structure inferred in target data. By contrast, supervised approach is more expressive, producing separable clusters in latent space that are coherent with topics, as shown in Figure~\ref{fig:cluster_vis3}-\ref{fig:cluster_vis4}. However, the latter model requires learning document embeddings with topics as label supervision. So, it is more costly than the unsupervised embedding approaches.

To negotiate the trade-offs between utilizing unsupervised and supervised approaches for learning document embeddings, we utilize keywords as sub-latent structure in corpora to train Seq2Seq networks referred to as \textbf{KeyGen2Vec}. Our work holds an assumption that learning a conditioned sequence-to-sequence mapping between documents and their corresponding keywords equals to learning the structural similarity that hierarchically links contents in document, keywords as explicit document abstractions, and topics as latent variables that further group documents based on keywords co-occurrences. For a fair comparison, we also train Multi-label and Multi-class Neural Network classifiers as supervised approaches to learn document embeddings on cQA data. The main difference between our proposal and classifier-based approaches is that the classifiers view keywords and topics as discrete labels $Y \in R^d$, while the proposed \textbf{KeyGen2Vec} sees keywords as a sequence of discrete structure $Y \in \Sigma^*$.

Summarizing, \textbf{our main contributions} are:
\begin{itemize}
    \item We introduce \textbf{KeyGen2Vec}, a simple Seq2Seq framework that can be utilized as a general tool to learn document embeddings conditioned on sub-topics information, such as keywords.

    \item We comprehensively investigate currently available approaches for learning document embeddings 
    
    We empirically show that unsupervised approaches often produce clusters that are incoherent with hidden semantics or latent structure inferred in target data.

    \item We empirically show that training Seq2Seq networks on multi-label keyword generation is analogous to indirectly incorporating label dependency assumption.

    We demonstrate that the proposed \textbf{KeyGen2Vec} is superior than a classifier that is trained on multi-label classification task with document source as inputs and keywords as target outputs for the models.
    

\end{itemize}


\section{Background}
\label{sec:backgr}

\subsection{Community-based Question Answering}

Our study focuses on investigating the potential usefulness of state-of-the-art document embeddings for clustering cQA archives with topics as latent structural similarity. Most of previous studies on cQA archives are centralized on the exploration of \textbf{retrieval} issues, such as learning latent topics for question retrieval \cite{cai-etal-2011-learning}, a retrieval framework with neural network embedding \cite{p-etal-2017-latent}, hybrid approach of neural network and latent topic clustering to rank the candidate answers given question \cite{yoon-etal-2018-learning}; and \textbf{textual similarity} problems between questions and their candidate answers \cite{wang-etal-2010-modeling,tan-etal-2016-improved,yang-etal-2018-learning}. Whereas, previous works on \textbf{clustering} cQA archives mainly focus on improving clustering algorithm based on simple feature extractor method (e.g. TfIdf) \cite{momtazi2009word,p-2016-mixkmeans}. \textbf{Topical clustering} itself is previously studied by \citet{rosa2011topical} to organize large unstructured twitter posts into topically coherent clusters with hashtags as a means of guidance.

\subsection{Document Embedding}
                  
\begin{table}[!ht]
    \centering
    \resizebox{.5\textwidth}{!}{
    \begin{tabular}{c | l |  c c c c c c c }
    \hline
    No & Model & GLO & SUB & SEQ & PRE & TRA & DIM \\
    \hline
    1 & \textbf{KeyGen2Vec} &-  & \checkmark & \checkmark & - & \checkmark & 200\\
    2 & S2S-AE &  - & -  & \checkmark & - & \checkmark& 200\\
    \cellcolor{Gray}3 & \cellcolor{Gray} FC-Mult-Cls $^{*)}$  & \cellcolor{Gray} \checkmark & \cellcolor{Gray} - & \cellcolor{Gray}  -  & \cellcolor{Gray} - & \cellcolor{Gray} \checkmark& \cellcolor{Gray} 100\\
    4 & Sigm-Mult-Lbl & - & \checkmark &  - & - & \checkmark& 100\\
    5 & Softm-Mult-Lbl &  - & \checkmark &  - & - & \checkmark& 100\\
    6 & S-BERT   & - & -  & - & \checkmark & - & 768 \\
    7 & LDA-Topic  & - & -  & - & - & \checkmark & $^*$  \\
    8 & D2V-DBOW100 & - & - &  - & - & \checkmark & 100\\
    9 & D2V-PVDM100 &  - & - & - & - & \checkmark & 100\\
    10 & Avg-GloVe100 & - & - &  - & \checkmark & - & 100\\
    11 & Avg-w2v100 & - & - & - & \checkmark & - & 100\\
    12 & Avg-GloVe300 &  - & - & - & \checkmark & - & 300\\
    13 & Avg-w2v300 &  - & - & - & \checkmark & - & 300\\
    14 & Avg-w2v50-tr-sm & - & - & - & - & \checkmark & 50\\
    15 & Avg-w2v50-tr-lg  & - & - & - & - & \checkmark & 50\\
    16 & Avg-PMI50 &  - & - & - & - & \checkmark & 50\\
    17 & DC-GloVe100 &  - & - &  - & \checkmark & - & $100^2/2$\\
    18 & DC-w2v100 & - & - &  - & \checkmark & - & $100^2/2$ \\
    19 & DC-PMI50 &  - & - &  - & - & \checkmark & $50^2/2$\\
    20 & DC-w2v50-tr-lg &  - & - &  - & - & \checkmark & $50^2/2$ \\
    21 & TfIdf &  - & -  &  - & - & \checkmark & $^{**}$\\
    \hline
    \end{tabular}
    }
    \caption{Properties of models compared in this study:  \textbf{GLO}: Global semantics (topic labels) are exposed to the model; \textbf{SUB}: Sub-semantic structures (keywords) are exposed to the model; \textbf{SEQ}: Model with sequential assumption; \textbf{PRE}: Use pretrained embedding - no finetuning; \textbf{TRA}: training on observed corpus; \textbf{DIM}: Dimension of embeddings; $^{*}$ depends on the chosen hyper-parameter ($N$ topics) ; $^{**}$ - depends on the vocabulary size in corpus. $^{*)}$ is upper bound model.}
    \label{tab:all_models}
\end{table}

Our study on currently available document embeddings is constrained on approaches that are domain independent. Since most space is devoted to the proposed framework and model evaluation, we refer the future readers to the original papers. Figure \ref{fig:doc_rep_app} shows document embedding approaches that are being observed in this study, which we broadly divided based on three categories: (1) Non-distributed (frequency-based) approach; (2) Probabilistic approach; and (3) Distributed (neural-based) embedding learning. The property of each embedding model is briefly described in Table~\ref{tab:all_models}. For a fair comparison, we include methods that learn embeddings based on global semantic structure (\textbf{GLO}), sub-semantic structure (\textbf{SUB}), sequential assumption (\textbf{SEQ}), pretrained embeddings (\textbf{PRE}), and directly trained embeddings on the target corpus (\textbf{TRA}).

\section{KeyGen2Vec Framework}
\label{sec:substruct}

\begin{figure}[!ht]
     \centering
	\includegraphics[width=\linewidth]{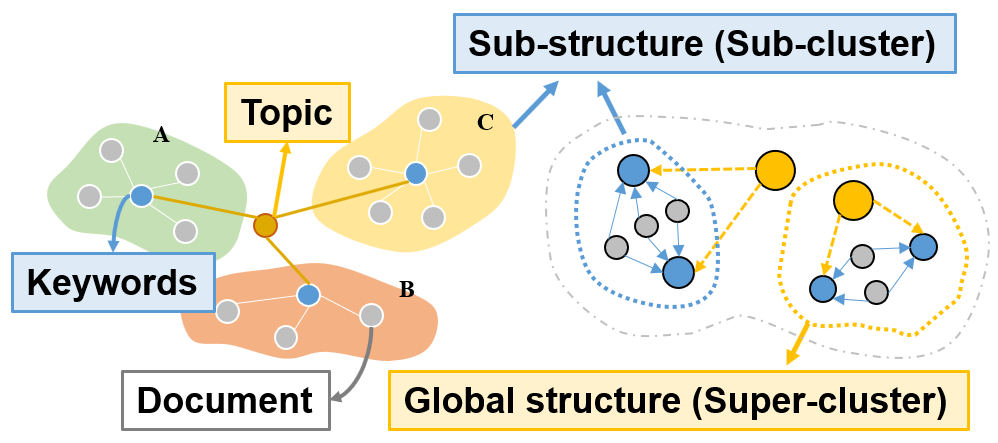}  
        \caption{Corpus as hierarchical semantic network of documents, keywords, and topics.}
        \label{fig:sub-struct-cluster}
\end{figure}

\textbf{KeyGen2Vec} is built based on a hierarchical semantic assumption of a corpus, as briefly illustrated in Figure~\ref{fig:sub-struct-cluster}. The assumption is that documents and their corresponding keyword labels form sub-structures or \emph{sub-networks} of latent topic structure as global semantics. Our work adopts Seq2Seq-based keyphrase generation introduced by \citet{meng-etal-2017-deep,chen-etal-2018-keyphrase}. While these preliminary works are motivated by the intuition of Seq2Seq capturing document semantics, there is currently neither analysis nor empirical evidence to support the claim that the learnt context representation has encapsulated latent semantic concept of document source conditioned on its keyword labels. We hypothesize that Seq2Seq network that has been trained on a keyword generation task is capable of capturing such latent semantic structure inferred in data.

\begin{figure}[!ht]
     \centering
	\begin{subfigure}[ht]{0.15\textwidth}
         \centering
         \includegraphics[width=\linewidth]{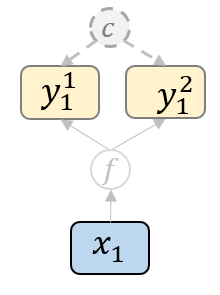}
	\caption{}
	\label{xy1}
     \end{subfigure}  
	 \begin{subfigure}[ht]{0.15\textwidth}
         \centering
         \includegraphics[width=\linewidth]{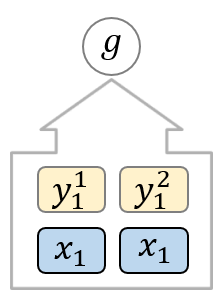}
	\caption{}
	\label{xy2}
       \end{subfigure}
     \begin{subfigure}[ht]{0.15\textwidth}
         \centering
         \includegraphics[width=\linewidth]{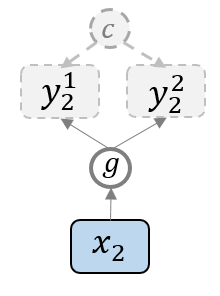}
	\caption{}
	\label{xy3}
       \end{subfigure}
	\caption{Training and Inference stages of the proposed KeyGen2Vec; (a) Observation set; (b) Training; (c) Prediction in test set; $c$ is latent topic.}
    \label{fig:train_xy}
    \vspace{-.5em}
\end{figure}

Figure \ref{fig:train_xy} illustrates the reformulation of multi-label keyword generation as the training objective of \textbf{KeyGen2Vec}. The objective of the task is to approximate the mapping function $f: X \mapsto Y$ - where $X$ denotes a collection of documents and $Y$ denotes the corresponding set of keywords in observation set. These sets of observations $\{(x_i, \{y_i^1, y_i^2\})\}$ were transformed into one-to-one training examples $\{x_i, y_i^1\}, \{x_i, y_i^k\}$ (fig. \ref{xy2}). Each training example is represented as sequences, $\mathcal{X}: \Sigma^*$ and $\mathcal{Y}: \Sigma^*$. In inference stage, to evaluate how well the trained Seq2Seq capture the semantic structure inferred in $f$, the parameterized encoder decoder model $g$ was further utilized as a decoder framework, to generate keywords given unseen documents. Details of architecture used is further explained in sec.\ref{sec:seq2seq_KeyGen2Vec}.

\subsection{Architecture}
\label{sec:seq2seq_KeyGen2Vec}

Our framework is built based on a standard Sequence-to-Sequence (Seq2Seq) encoder-decoder framework. An encoder first maps a sequence of words to a vector $c$ -- where $c$ serves as the resulting document embedding. Given the encoded embedding of document source $c$, the decoder then generates target sequences. 

\vspace{-1.5em}
\begin{align*}
    \texttt{ENC:} x = \{w_1, \ldots, w_{T_x} \} \mapsto c \in \mathbb{R}^d \\
    \texttt{DEC:} c \in \mathbb{R}^d \mapsto  y = \{w_1, \ldots, w_{T_y} \}
    \label{eq:enc-dec}
\end{align*}

\paragraph{Encoder}

The encoder network is constructed of bidirectional GRU units for mapping sequence of embedded words $e_{t \cdots T_x}$ into a sequence of intermediate state representation $h_{t \cdots T_x}$, which is a concatenation of forward and backward hidden states $h_{t \cdots T_x} = [\overrightarrow{h}, \overleftarrow{h}]$. 

\begin{align*}
    \overrightarrow{h_{t \ldots T_x}} = \overrightarrow{GRU} (e_{t \cdots T_x}) \\
    \overleftarrow{h_{t \ldots T_x}} = \overleftarrow{GRU} (e_{t \cdots T_x})
\end{align*}

\paragraph{Decoder}

The decoder is a neural language model based on forward GRU network that conditions on context embedding of encoder $c$. $s_{t-1}$ is decoder state at previous time step. $y_{t-1}$ denotes prediction at $t-1$. Here, $g(.)$ denotes prediction layer (dense network) with softmax activation function.

\begin{align*}
        s_t = \overrightarrow{\rm GRU}(y_{t-1}, s_{t-1}, c) \\
        p(y_t|y_{1, \ldots, t-1}, x) = g(y_{t-1}, s_t, c)
\end{align*}

\paragraph{Attention}

We use Bahdanau's MLP attention scoring function \cite{Bahdanau2014Neural} to calculate attention score $\alpha$ corresponds to the importance weight of words in source sequence given embedding of words in target sequence.
\begin{align*}
    \alpha_{t} = \frac{exp(score(s_t^{(j)},h_{t \cdots T_x}^{(i)}))}{\sum_t^{T_x}{exp(score(s_t^{(j)},h_{t \cdots T_x}^{(i)}))}}
\end{align*}

\paragraph{Context (Document) Embeddings}

The final document embedding $c$ is computed based on weighted sum between a sequence of encoder states and attention score.
\vspace{-1em}
\begin{align*}
    {c}_{t} = \sum_{t=1}^{T_x}{\alpha_{t} h_{t}}
\end{align*}

\subsection{On Label Dependency Assumption} 
\label{sec:dependency}




In our proposed \textbf{KeyGen2Vec} framework, keywords as target variables are represented as sequences of words. The probability of a particular keyword chosen in inference stage equals to the joint probability of words in sequence $p(w_{1:T})$. Softmax activation function is used for projecting decoder states $s_{t-1} \in R^d$ into probabilistic values over $\mathcal{V}$ vocabulary size, $\in R^{\mathcal{V}}$. 
\vspace{-.5em}
\begin{align*}
        p(w_t | w_1, \ldots, w_{t-1};\theta) = \texttt{\small softmax}(W s_{t-1} + b) \\
        p(w_{1:T}) = \prod_{t} p(w_t | w_1, \ldots, w_{t-1})
\end{align*}

\noindent where softmax function is formally given by:
\vspace{-.5em}
\begin{align*}
        \texttt{\small softmax}(z)_i = \frac{e^{z_i}}{\sum_{j=1}^K e^{z_j}} 
\end{align*}

By dividing each softmax unit (the probability of each word $w_t$ in vocabulary $\mathcal{V}$) with the sum of all units, the total probability of words in $\mathcal{V}$ is ensured to be $1$. An increase of one class probability $p (y_t|x, \theta)$ causes the probability of other class decreases. 

We hypothesize that by transforming one-to-many training objective in multi-label keyword generation task into one-to-one multi-class learning scheme, as shown in Figure~\ref{fig:train_xy}), we indirectly incorporate label dependency assumption during training stage. The trained model treats each sample as mutually exclusive event via softmax normalization and outputs final prediction $\hat{y_t} = \argmax p (y_t|x, \theta)$. This results in an indirect dependent assumption between a pair of keyword labels since the probability of particular pair of keywords given the same document source $p(y_1^1|x_1)$ and $p(y_1^2|x_1)$ are dependent each other. By contrast, standard multi-label learning commonly uses independent Bernoulli assumption via Sigmoid function, disregarding the dependency between labels. We further investigate this problem by comparing models with Softmax-based multi-class classification loss and a standard Sigmoid-based Multi-label classifier.

\section{Experiments}

\subsection{Data}

\label{sec:datasets}
We use the following data constructed from cQA archives as gold standard for learning and evaluation. The three data sets represent data with different level of difficulties w.r.t. sentence length, noise-level, and number of unique keywords and topic labels. Toy data is considered to be less noisy and balance -- each sub-class category is composed of sentences and their paraphrases, forming natural cluster structure. Yahoo! data sets with 5 topic categories (5-T) and 11 topics (11-T) are considered to be more noisy and imbalanced due to many non-informative words (e.g. digits, measures, url-links, query about address or web sources) and domain specific terms (e.g. medical and automotive terms).

\begin{table}[!ht]
    \centering
    \resizebox{.47\textwidth}{!}{
    \begin{tabular}{l|c|c|c|c|c}
    \hline
    Data set & \#Topics  & \#Keywords  & \#Train & \#Test & Sentence\\
    & (GLO) & (SUB) &&& Length\\
    \hline
        Wikianswer & NA & NA & ~700M & NA & 9 $\pm$ 3 \\
        Toy data & 12 & 77 & 1158 & 290 & 9 $\pm$ 3\\
        5-T Yahoo! cQA & 5 & 120 & 23824 & 5957 & 36 $\pm$ 28\\
        11-T Yahoo! cQA & 11 & 179 & 70962 & 17741 & 35 $\pm$ 28 \\
    \hline
    \end{tabular}
    }
    \caption{Data set; Wikianswer (original corpus of Toy data) is used to train Word2Vec (w2v50-tr-lg) and PMI method.}
    \label{tab:dataset}
\end{table}

\vspace{-1em}
\paragraph{Toy Data}
We created a small set of hand-labelled sentence-keywords-topic pairs (1448 sentences) from WikiAnswer \footnote{http://knowitall.cs.washington.edu/oqa/data/wikianswers/}. WikiAnswer is a data set composed of millions of questions asked by humans, where each sentence example is accompanied by its paraphrased versions, forming a paraphrase cluster of one particular question. We use the original WikiAnswer corpus to train large scale Word2Vec and PMI models incrementally, to inspect how the scale of data affects model performance. Table. \ref{tab:toy_examples} shows a training example in Toy data. The number of keywords and topic assignment per sentence were made fixed, i.e. two keywords and one topic for each sentence.

\begin{table}[!ht]
    \centering
    \begin{tabular}{| p{7cm} |}
    \hline
    \vfill
    \textbf{Source}: ``the sporozoan plasmodium carried from host to host by mosquitoes causes what serious infection?\\
    \vfill
    \textbf{Keywords:} malaria; plasmodium parasite \\
    \textbf{Topic:} virus and diseases\\ 
    \hline
    \end{tabular}
    \caption{Sentence examples in Toy Data set}
    \label{tab:toy_examples}
\end{table}

\begin{table}[!ht]
    \centering
    \begin{tabular}{ |p{7cm}| }
    \hline
    \vfill
    \textbf{Source}: ``what is diabetes mellitus? diabetes mellitus is medical disorder characterized by varying or persistent hyperglycemia elevated blood sugar levels, $\dots$''  \\
    \vfill
    \textbf{Keywords:} diabetes; diseases and conditions \\
    \textbf{Topic:} health \\ 
    \hline
    \end{tabular}
    \caption{Sentence example (concatenated cQA pair) in Yahoo! Answer cQA}
    \label{tab:qa_yahoo_examples1}
\end{table}

\paragraph{Yahoo! Answer Comprehensive cQA}

We reproduce and extend our result on real world cQA archives consisting of question-answers pairs, accompanied by keywords (tags) and the corresponding topic. Data was obtained from Yahoo! Answer Comprehensive cQA dataset  \footnote{https://webscope.sandbox.yahoo.com/catalog.php}, originated from the query log of Yahoo! Answer. We constructed two corpora: corpus with 5 topic categorization -- referred to as 5-T cQA and corpus with 11 topics -- referred to as 11-T cQA. The training and test examples were constructed by concatenating each question and the corresponding answers. Table \ref{tab:qa_yahoo_examples1} shows a training example obtained from Yahoo! Answer cQA archives. Likewise, each document corresponds to a fixed membership: two keywords and one topic describing the document semantic abstraction. 

\subsection{Training and Hyper-parameters}

For training \textbf{KeyGen2Vec}, we use negative log-likelihood loss function with an adaptive learning rate optimization (Adam \cite{kingma2014adam}), $lr=0.001, betas=(0.9, 0.98), eps=1e-9$. Curriculum learning \cite{Bengio2015Scheduled} was employed to sampling whether to use a teacher forcing method during training stage. For the other models, we refer the reader to the provided code documentation.

For LDA, trainable Word2Vec, Paragraph Vector, we used Gensim implementation \footnote{https://radimrehurek.com/gensim/}. BERT pretrained sentence encoder is taken from a recent sentence similarity task \cite{reimers-gurevych-2019-sentence}. Specific for Toy data experiment, we trained two Word2Vec models: small scale model \texttt{\small Avg-w2v50-tr-sm} was trained on the constructed set of Toy data; and large scale model \texttt{\small Avg-w2v50-tr-lg} was trained incrementally on WikiAnswer (the original large scale corpora of Toy data) -- to inspect how model performance differs based on the scale of data. Classifiers (Multi-class and Multi-label) were constructed from fully-connected network (MLP) since we do not find a significant performance differences between using different types of networks (i.e. dense, convolutional, and recurrent). We trained two types of Multi-label classifiers (\texttt{\small Sigm-Cls} and \texttt{\small Softm-Cls}) to inspect the effect of incorporating label dependency in multi-label learning.  

\subsection{Clustering as Evaluation}

We use K-Means clustering \footnote{scikit-learn.org/../sklearn.cluster.KMeans.html} to evaluate the quality (clusterability) of document embeddings in this study (table \ref{tab:all_models}). The hyper-parameter choices of K-means is kept as minimum as possible (\texttt{\small init='random', n\_clusters=$K$, n\_init=$10$, max\_iter=$50$}). This is to make sure that the clustering is not overly parameterized, which can obscure the actual quality of the learnt embeddings. Given the actual global semantic classes (topic labels) $\mathbb{C}$ in the current observed corpora and the predicted classes $\Omega$ from K-Means method, we employ \textbf{Purity}, \textbf{Normalized Mutual Information (NMI)}, and \textbf{F1-score} \cite{manning2008introduction} metrics to objectively measure whether the resulting clustering $\Omega$ can recreate or approximate the exact classes $\mathbb{C}$.

\subsection{$\chi^2$ Feature selection}
\label{sec:chi2-feats}

\begin{table}[!ht]
    \centering
    \resizebox{.45\textwidth}{!}{
    \begin{tabular}{ | c | c | c | c | }
    \hline
     \textbf{Dining Out} & \textbf{Health} & \textbf{Travel} & \textbf{Cars} \\
     \hline
     hamburger & medicine & trip & jeep \\ 
     taco & heart & map & vehicle \\ 
     sandwich & symptom & disney & auto \\ 
     buffet & virus & vacation & manual\\ 
     cafe & treatment & ticket & nisan \\ 
     \hline
    \end{tabular}
    }
    \caption{Example of influential words per topic category -- selected based on $\chi^2$ feature selection method on 5-T Yahoo! cQA.}
    \label{tab:chi-1b}
\end{table}


We employ feature selection based on $\chi^2$ method \cite{manning2008introduction} to select $N-$ most influential words per topic category. Each training example is then represented as Bag-of-Influential words with $N \in \{20,50,100,250\}$ for Toy data and $N \in \{20,50,100,250,500,1000,2500\}$ for Yahoo! cQA data. The larger the size of influential words per category, the more noises preserved in the training data. This experiment was conducted to investigate: (1) the effect of noises on the clusterability of embeddings; (2) the effect of incorporating label dependency via Softmax-based loss on the clusterability of embeddings.

\begin{figure*}[!t]
    \centering
     \begin{subfigure}[ht]{0.15\textwidth}
         \centering
         \includegraphics[width=\linewidth]{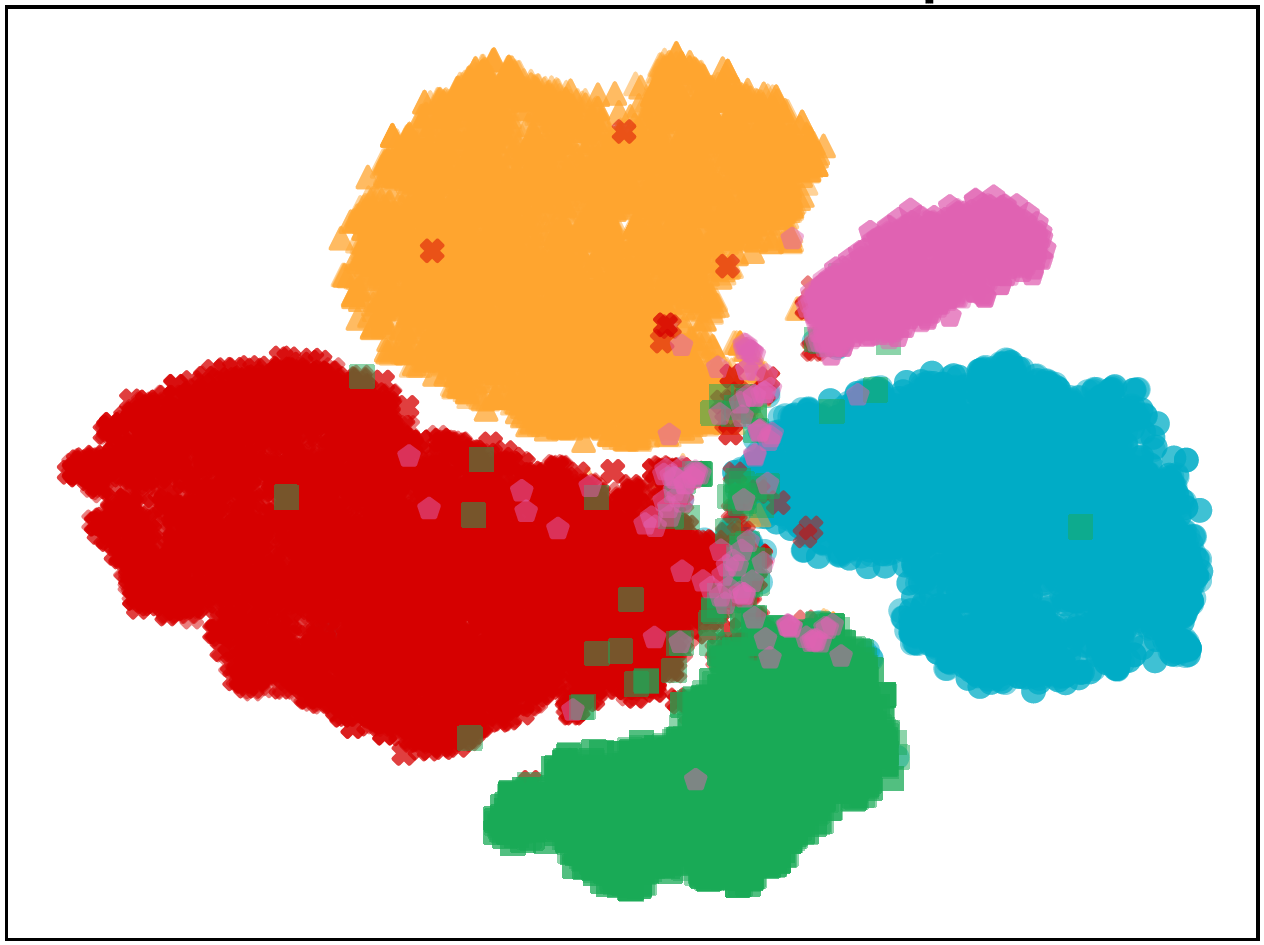}
         \caption{Upper-bound}
         \label{fig:mc-true}
     \end{subfigure}
     \begin{subfigure}[ht]{0.15\textwidth}
         \centering
         \includegraphics[width=\linewidth]{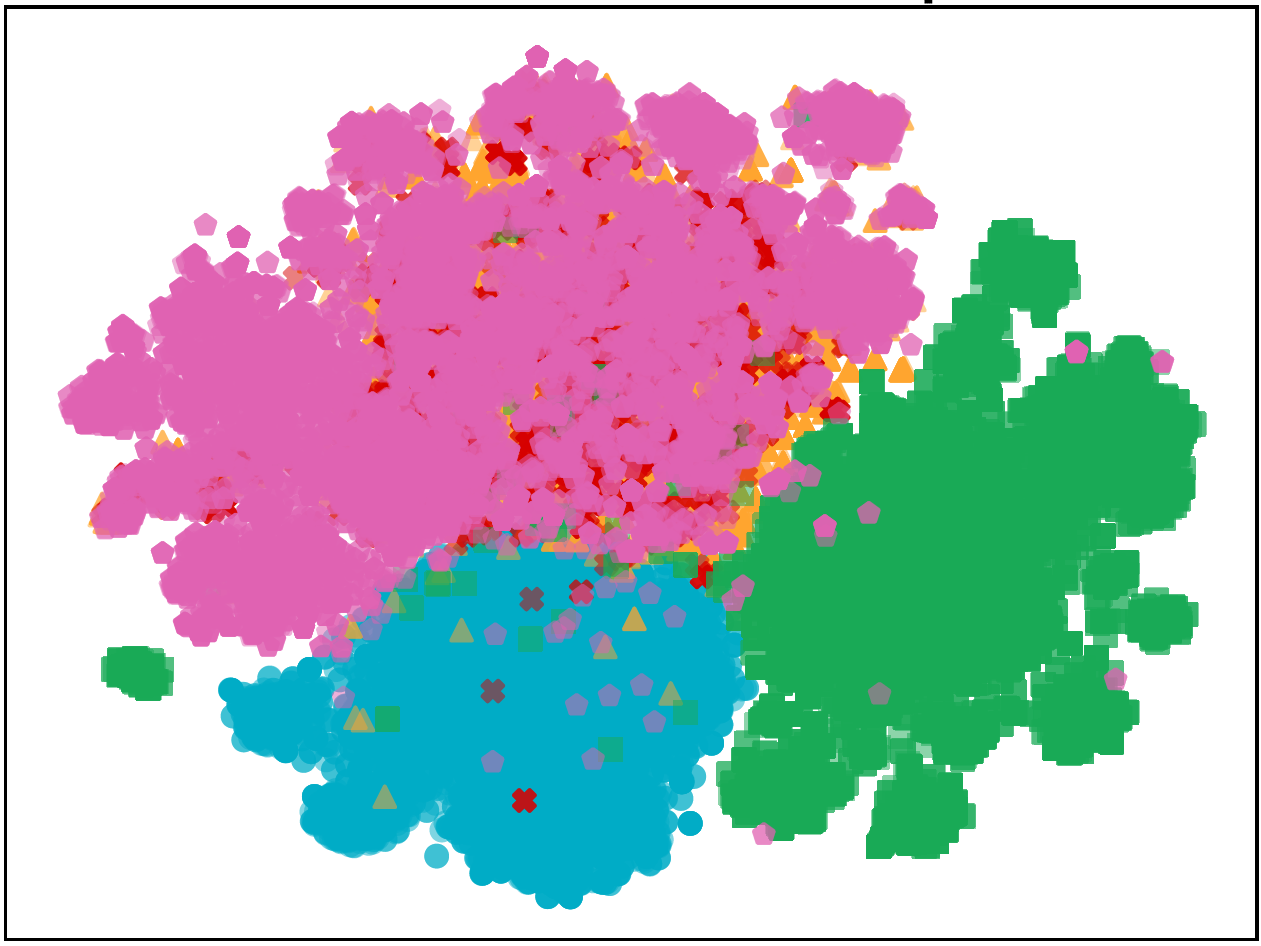}
         \caption{KeyGen2Vec}
         \label{fig:key-true}
     \end{subfigure}
     \begin{subfigure}[ht]{0.15\textwidth}
         \centering
         \includegraphics[width=\linewidth]{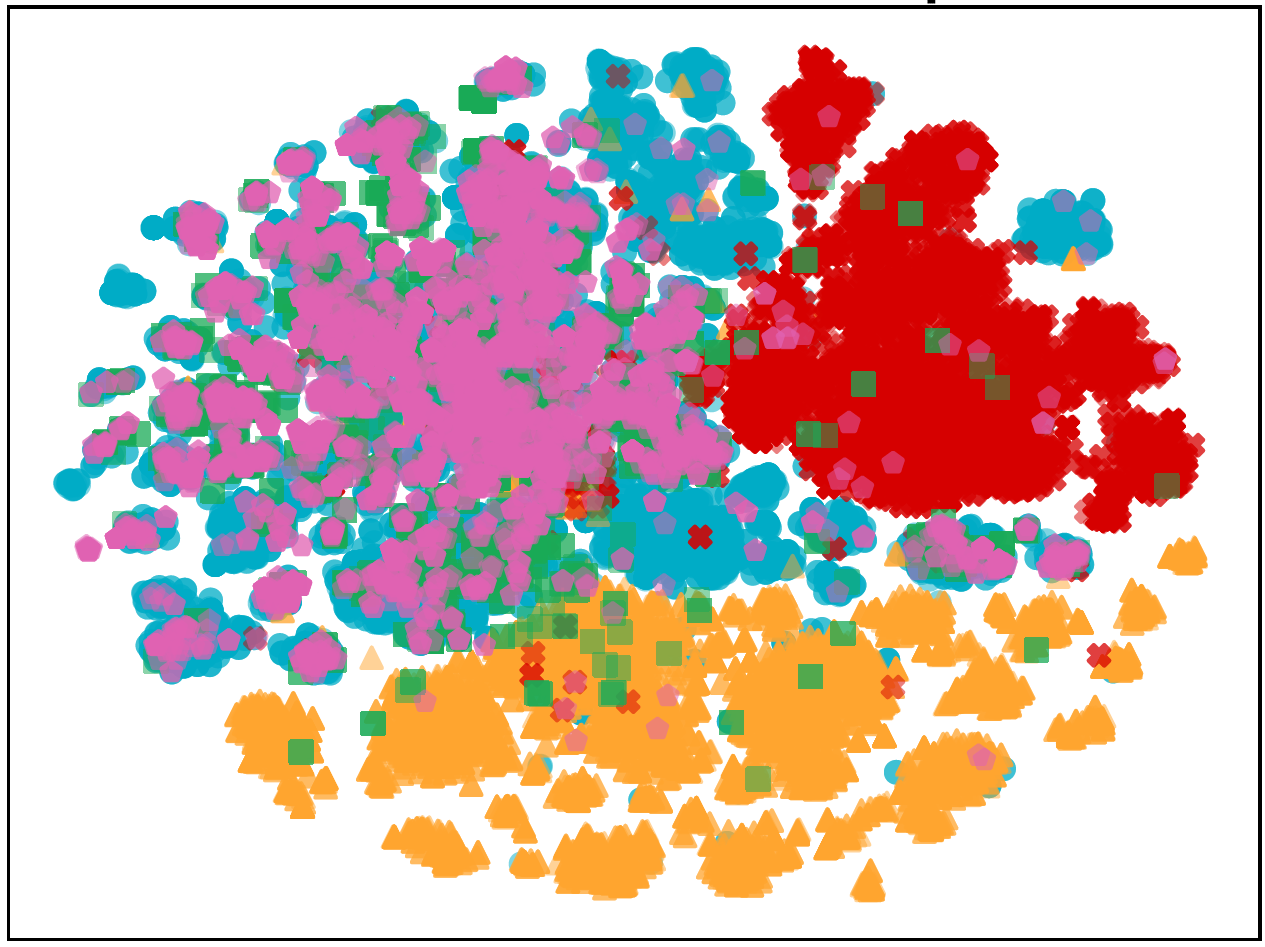}
         \caption{Sigm-MultLbl}
         \label{fig:sigm-true}
     \end{subfigure}
     \begin{subfigure}[ht]{0.15\textwidth}
         \centering
         \includegraphics[width=\linewidth]{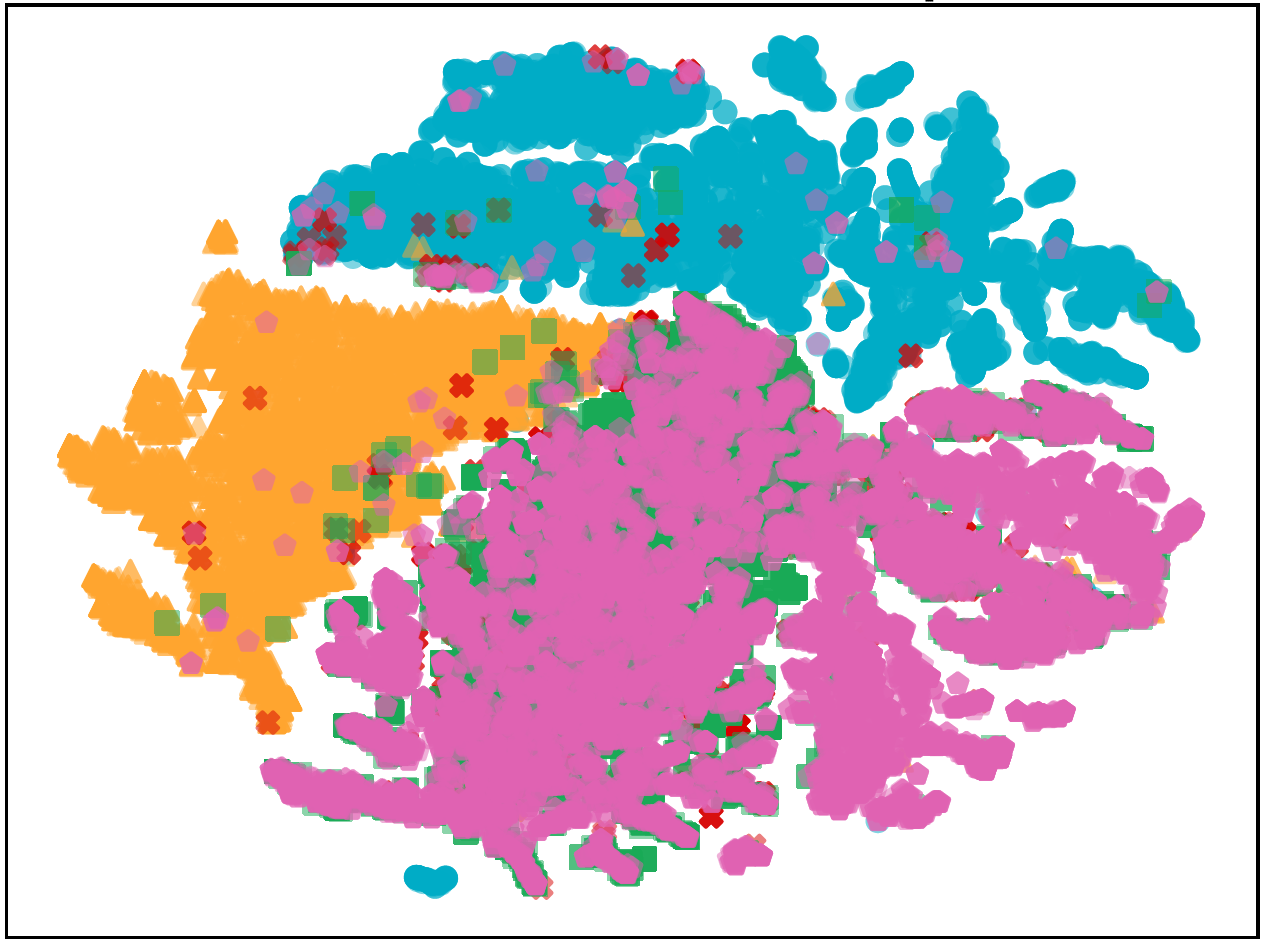}
         \caption{Softm-MultLbl}
         \label{fig:softm-true}
     \end{subfigure}
     \begin{subfigure}[ht]{0.15\textwidth}
         \centering
         \includegraphics[width=\linewidth]{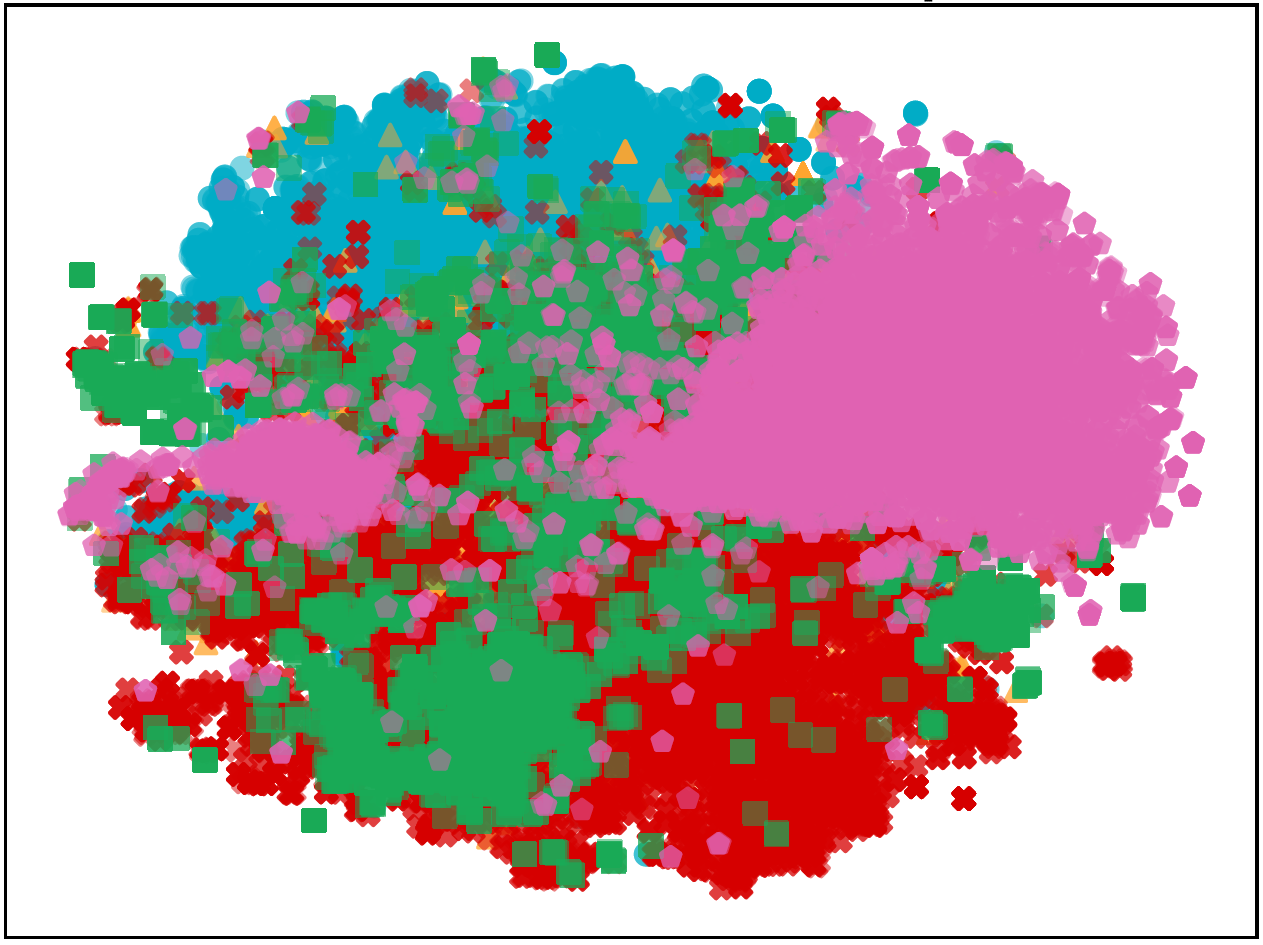}
         \caption{Avg-w2v-tr}
         \label{fig:w2vtr-true}
     \end{subfigure}
     \caption{Example of Clustering Visualization on 5-T Yahoo! Q\&A. }
	\label{fig:cluster-1b}
\end{figure*}

\section{Results and Discussion}
\label{sec:rsl_analyze}


We summarize our empirical findings as follows:

\paragraph{KeyGen2Vec outperforms multi-label classifiers}

Based on the clustering performance on three data sets, as shown in Table~\ref{tab:cluster_toy}-\ref{tab:cluster_11t_qa}, we demonstrate that although the model does not exploit the actual topic labels during training stage, the proposed \textbf{KeyGen2Vec} has a capability of preserving topical proximity in latent space, outperforming its counterparts -- models trained on multi-label classifiers (\texttt{\small Sigm-Mult-Lbl} and \texttt{\small Softm-Mult-Lbl}).

\begin{table}[!ht]
  \resizebox{.5\textwidth}{!}{
    \centering
    \begin{tabular}{l | c c | c c  | c c }
    \hline
    Approach & \multicolumn{2}{c}{Purity} & \multicolumn{2}{c}{NMI} & \multicolumn{2}{c}{F$_1$-score} \\
     & All & Test & All & Test &  All & Test \\
    \hline
    \textbf{KeyGen2Vec} & \bf  0.734 & \bf 0.726 & \bf 0.769 & \bf 0.774 & \bf 0.631 & \bf 0.614  \\
    S2S-AE & 0.288 & 0.317 & 0.213 & 0.269 & 0.167 & 0.164  \\
    \cellcolor{Gray}FC-Mult-Cls$^*$ & \cellcolor{Gray} 0.961 & \cellcolor{Gray} 0.938 & \cellcolor{Gray} 0.968 & \cellcolor{Gray} 0.940 & \cellcolor{Gray} 0.947 & \cellcolor{Gray} 0.910  \\
    Sigm-Mult-Lbl & 0.590 & 0.605 & 0.629 & 0.649 & 0.435 & 0.438 \\
    Softm-Mult-Lbl & 0.659 & 0.687 & 0.673 & 0.710 & 0.532  & 0.556 \\
    BERT & 0.636 & 0.644 & 0.680 & 0.678 & 0.527 & 0.502 2 \\
    LDA-Topic & 0.474 & 0.551 & 0.472 & 0.584 & 0.368 & 0.454  \\
    D2V-DBOW100 & 0.179 & 0.207 & 0.049 & 0.115 & 0.111 & 0.100 \\
    D2V-PVDM100 & 0.171 & 0.206 & 0.035 & 0.104 & 0.102 & 0.094  \\
    Avg-GloVe100 & 0.648 & 0.649 & 0.656 & 0.672 & 0.507 & 0.485 \\
    Avg-W2V100 & 0.686 & 0.679 & 0.694 & 0.689 & 0.535 & 0.489 \\
    Avg-GloVe300 & 0.668 & 0.693 & 0.688 & 0.719 & 0.523 & 0.531  \\
    Avg-W2V300 & 0.655 & 0.654 & 0.685 & 0.688 & 0.511 & 0.475 \\
    Avg-W2v50-tr-sm & 0.208 & 0.243 & 0.076 & 0.152 & 0.116 & 0.112 \\
    Avg-W2v50-tr-lg & 0.643 & 0.651 & 0.671 & 0.673 & 0.538 & 0.506 \\
    Avg-PMI50 & 0.305 & 0.325 & 0.212 & 0.290 & 0.180 & 0.176 \\
   DC-GloVe100 & 0.374 & 0.326 & 0.351 & 0.321 & 0.222  & 0.185 \\
    DC-W2V100 &  0.407 & 0.312 & 0.447 & 0.325 & 0.225 & 0.180 \\
    DC-PMI50 & 0.282 & 0.313 & 0.180 & 0.266 & 0.162 & 0.167  \\
    DC-W2V50-tr-lg & 0.555 & 0.546 & 0.545 & 0.567 & 0.376 & 0.359 \\
    TfIdf &  0.564 &0.611 & 0.615 & 0.649 & 0.377 & 0.401   \\
    \hline
    \end{tabular}
    }
     \caption{Clustering Evaluation on Toy Data. FC-Mult-Cls$^*$ is an upper bound model. Scores were calculated based on average score in 10 iterations of K-means clustering. \textbf{Higher} is \textbf{better}.}
    \label{tab:cluster_toy}
\end{table}

\begin{table}[!ht]
\resizebox{.5\textwidth}{!}{
    \centering
    \begin{tabular}{l | c c | c c  | c c }
    \hline
    Approach & \multicolumn{2}{c}{Purity} & \multicolumn{2}{c}{NMI} & \multicolumn{2}{c}{F$_1$-score} \\
     & All & Test & All & Test &  All & Test \\
    \hline
    \textbf{KeyGen2Vec} & \bf 0.801 &\bf  0.784 &\bf  0.657 & \bf 0.603 & \bf  0.668 & \bf  0.630 \\
    S2S-AE & 0.337 &  0.341 & 0.012 & 0.014 & 0.241 & 0.242 \\
    \cellcolor{Gray}FC-Mult-Cls$^*$ & \cellcolor{Gray} 0.970 & \cellcolor{Gray} 0.853 & \cellcolor{Gray} 0.903 & \cellcolor{Gray} 0.721 & \cellcolor{Gray} 0.957 & \cellcolor{Gray} 0.794 \\
    Sigm-Mult-Lbl & 0.763 & 0.738 & 0.567 & 0.522 & 0.635 & 0.606  \\
    Softm-Mult-Lbl & 0.772 & 0.744 & 0.579 & 0.533 & 0.657 & 0.622 \\
    BERT & 0.329 & 0.325 & 0.005 & 0.007 & 0.278 & 0.283\\ 
    LDA-Topic & 0.541  & 0.583 & 0.222 & 0.269 & 0.479 & 0.488 \\
    D2V-DBOW100 & 0.335 &  0.335 & 0.018 & 0.019 & 0.245 & 0.246 \\
    D2V-PVDM100 & 0.337 & 0.339 & 0.012 & 0.012 & 0.274 & 0.276 \\
    Avg-GloVe100 & 0.495 & 0.502 & 0.122 & 0.133 & 0.330 & 0.349\\
    Avg-W2V100 & 0.531 & 0.533 & 0.202 & 0.207 & 0.371 & 0.379 \\
    Avg-GloVe300 & 0.483 & 0.498 & 0.117 & 0.132 & 0.334 & 0.352  \\
    Avg-W2V300 & 0.463 & 0.462 & 0.117 & 0.117 & 0.342 & 0.347  \\
    Avg-W2V50-tr & 0.609 & 0.621 & 0.318 & 0.327 & 0.477 & 0.491  \\
    Avg-PMI50 & 0.339 & 0.343 & 0.024 & 0.026 & 0.276 & 0.273  \\
   DC-GloVe100 & 0.359 & 0.361 & 0.031 & 0.034 & 0.296 & 0.311  \\
    DC-W2V100 & 0.426 & 0.413 & 0.117 & 0.105 & 0.356 & 0.353 \\
    DC-PMI50 & 0.326 & 0.329 & 0.018 & 0.019 & 0.295 & 0.292 \\
    DC-W2V50-tr & 0.329 & 0.328 & 0.012 & 0.013 & 0.343 & 0.342  \\
    TfIdf & 0.357 & 0.383 & 0.047 & 0.077 & 0.304 & 0.308 \\
    \hline
    \end{tabular}
    }
    \caption{Clustering Evaluation on 5-T Yahoo! cQA. }
    \label{tab:cluster_5t_qa}
\end{table}

\begin{table}[!ht]
\resizebox{.5\textwidth}{!}{
\centering
    \begin{tabular}{l | c c | c c  | c c }
    \hline
    Approach & \multicolumn{2}{c}{Purity} & \multicolumn{2}{c}{NMI} & \multicolumn{2}{c}{F$_1$-score} \\
     & All & Test & All & Test &  All & Test \\
    \hline
    \textbf{KeyGen2Vec} & \bf 0.841 & \bf 0.797 & \bf 0.723 & \bf 0.662 & \bf 0.655 & \bf 0.603 \\
    S2S-AE & 0.306 & 0.303 & 0.037 & 0.036 & 0.135 & 0.133  \\
    \cellcolor{Gray}FC-Mult-Cls$^*$ & \cellcolor{Gray} 0.862 & \cellcolor{Gray} 0.768 & \cellcolor{Gray} 0.774 & \cellcolor{Gray} 0.643 & \cellcolor{Gray} 0.717 & \cellcolor{Gray} 0.564\\
    Sigm-Mult-Lbl & 0.729 & 0.723 & 0.545 & 0.535 & 0.487 & 0.479 \\
    Softm-Mult-Lbl & 0.739  & 0.718 & 0.589 & 0.538 & 0.508 & 0.493 \\
    BERT & 0.274 & 0.229 & 0.016 & 0.016 & 0.211 & 0.229  \\
    LDA-Topic & 0.518 & 0.534 & 0.275 & 0.300 & 0.304 & 0.293  \\
    D2V-DBOW100 & 0.279 & 0.278 & 0.022 & 0.024 & 0.133 & 0.133  \\
    D2V-PVDM100 & 0.281 & 0.280 & 0.019 & 0.018 & 0.167 & 0.162 \\
    Avg-GloVe100 & 0.435  & 0.434 & 0.193 & 0.199 & 0.212 & 0.214 \\
    Avg-W2V100 & 0.407 & 0.409 & 0.155 & 0.159 & 0.198 & 0.201 \\
    Avg-GloVe300 & 0.427 & 0.425 & 0.188 & 0.190 & 0.213 & 0.205 \\
    Avg-W2V300 & 0.448 & 0.449 & 0.216 & 0.221 & 0.216 & 0.218 \\
    Avg-W2V50-tr & 0.551 & 0.552 & 0.309 & 0.311 & 0.294 & 0.287  \\
    Avg-PMI50 & 0.283 & 0.283 & 0.032 & 0.035 & 0.138 & 0.139 \\
   DC-GloVe100 & 0.303 & 0.307 & 0.053 & 0.059 & 0.169 & 0.182 \\
    DC-W2V100 & 0.328 & 0.321 & 0.088 & 0.089 & 0.203 & 0.209  \\
    DC-PMI50 & 0.273 & 0.271 & 0.018 & 0.019 & 0.161 & 0.160 \\
    DC-W2V50-tr & 0.303 & 0.301 & 0.043 & 0.043 & 0.203 & 0.203 \\
    TfIdf & 0.306 & 0.303 & 0.037 & 0.037 & 0.135 & 0.133  \\
    \hline
    \end{tabular}
    }
    \caption{Clustering Evaluation on 11-T Yahoo! cQA. }
    \label{tab:cluster_11t_qa}
    \vspace{-.5em}
\end{table}

\paragraph{Pretrained unsupervised embeddings are more random on noisy datasets}
Specific to unsupervised pretrained models (S-BERT, Wikipedia2Vec \texttt{\small Avg-w2v100} and GloVe \texttt{\small Avg-GloVe300}), the results show that while these models perform well in Toy data, their performance degrades in the other two data sets. This indicates that the embeddings generalized from the source domain in which the model is trained on are not sufficient for the target corpora (Yahoo! data). Fine tuning the models or expanding the vocabulary, however, is beyond the scope of this study.  

This finding specifically challenges the prior belief stating an \emph{off-the-shell} encoder that has been trained on large scale data or multi-tasks learning (e.g. Skip-Thought vector, Universal Sentence Encoder, Sentence-BERT, pretrained word embeddings) can produce highly generic embedding that performs well in practice. We argue that a generic pretrained embedding may best fit for tasks that are less noisy and complement to the pretrained source domain, exemplified in our Toy data experiment.


\paragraph{Trained unsupervised embeddings rely on large-scale data}
We observe that unsupervised neural embeddings that are trained on the observed corpora (e.g. Word2Vec, Doc2Vec, Seq2Seq Autoencoder) seemingly rely on the scale of data. See how small scale Word2Vec (\texttt{\small Avg-w2v50-tr-sm}) results in a notably low performance on Toy data (similar to Autoencoder \texttt{\small S2S-AE} and Doc2Vec), as compared to large scale Word2Vec (\texttt{\small Avg-w2v50-tr-lg}).

\begin{figure}[!ht]
     \centering
	\begin{subfigure}[ht]{\linewidth}
         \centering
         \includegraphics[width=\linewidth]{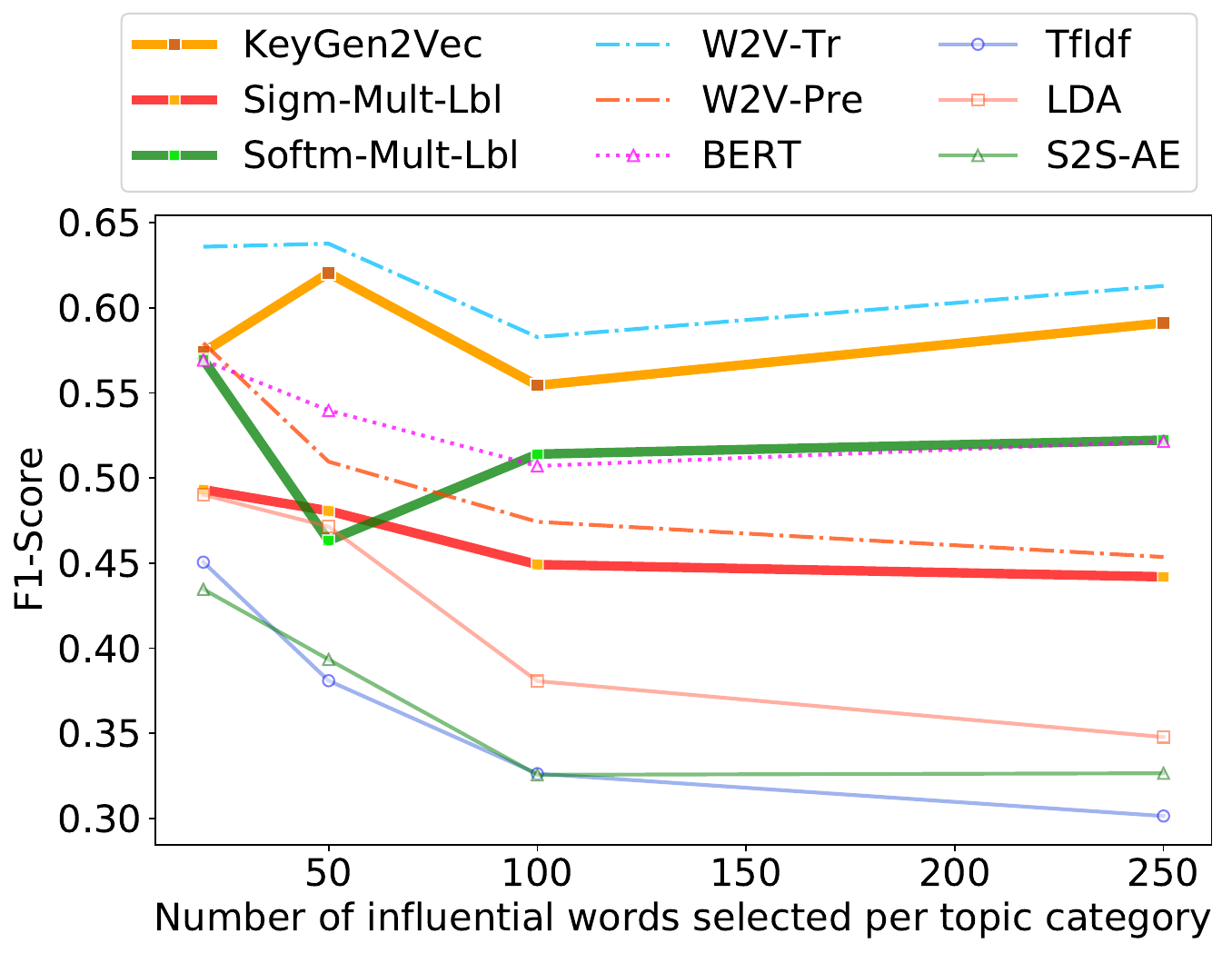}
	\caption{Toy Data}
	\label{fpr:toy}
        \end{subfigure}   
	\begin{subfigure}[ht]{\linewidth}
         \centering
         \includegraphics[width=\linewidth]{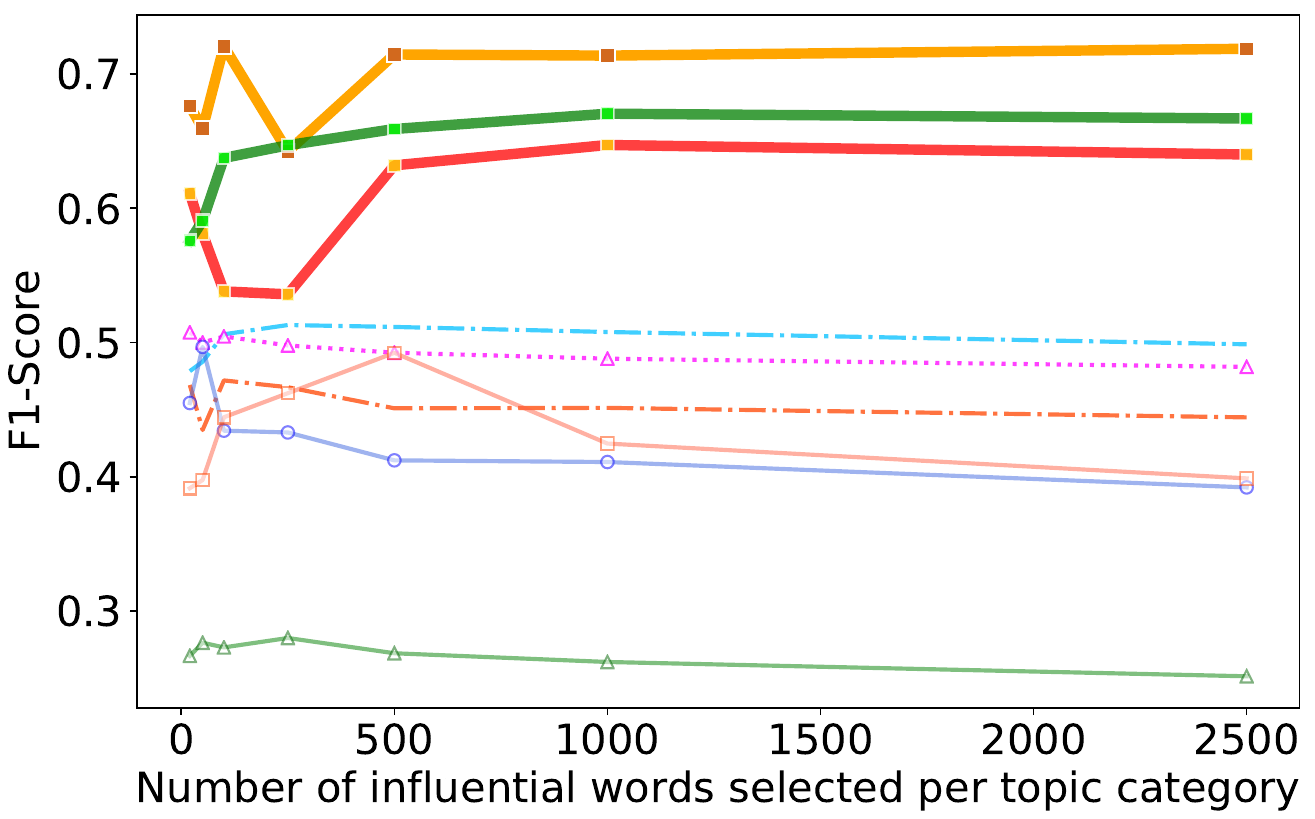}
	\caption{5-T Yahoo! Answer Data}
	\label{fpr:5t}
        \end{subfigure} 
    \begin{subfigure}[ht]{\linewidth}
         \centering
         \includegraphics[width=\linewidth]{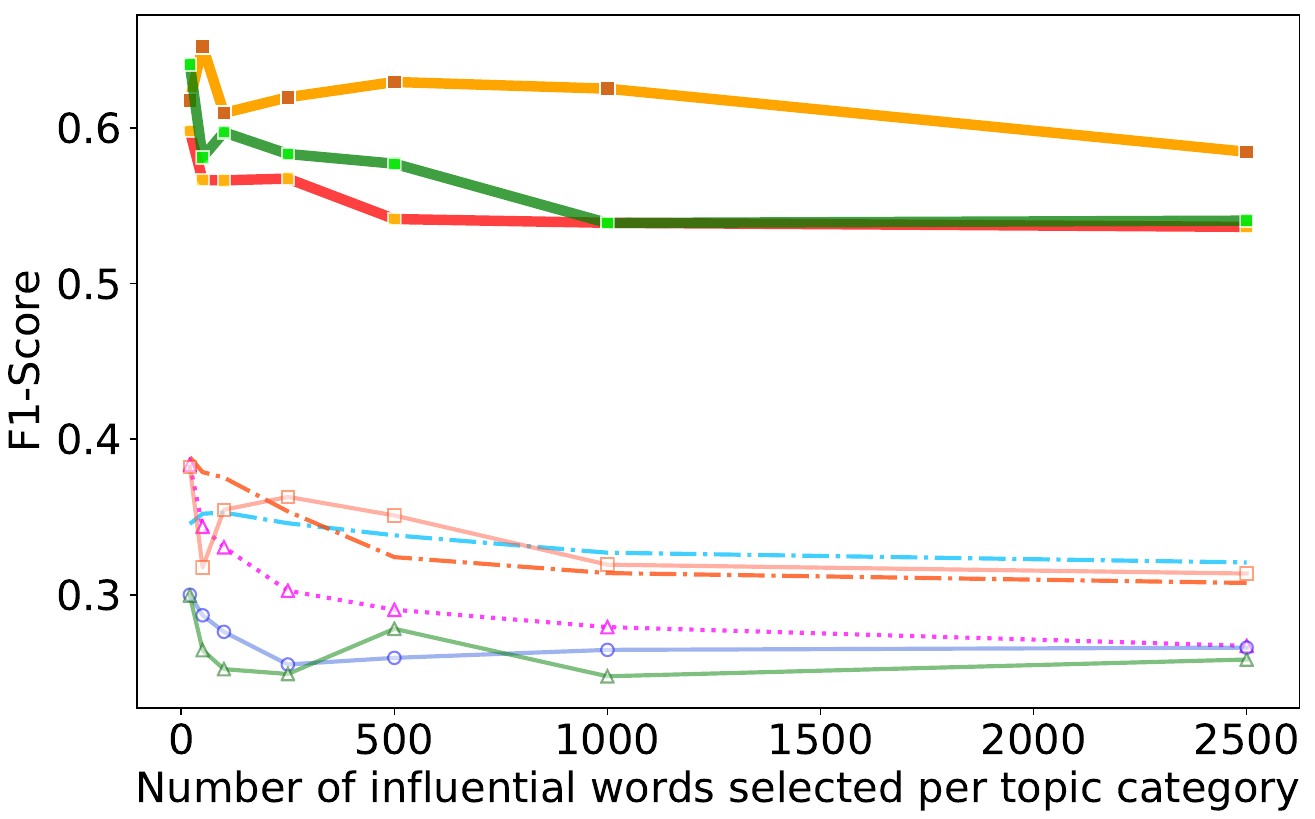}
	\caption{11-T Yahoo! Answer Data}
	\label{fpr:11t}
        \end{subfigure} 
        \caption{Effect of noises on F1-Score; \emph{best viewed in color. The \textbf{larger} pre-selected feature set size (x-axis), the \textbf{more noisy} the documents are.}}
        \label{fig:effect-fpr}
        \vspace{-.5em}
\end{figure}

We argue that the low quality of unsupervised embeddings in the current study is due to the models mainly depend on local information in document contents --  there is no strong assumption on differentiating salient features (words) w.r.t. global semantic aspects, which may hinder their direct utilization on a subsequent predictive analytics tasks. Specific to LDA topic model, we argue that their low performance in the current task is due to no strong assumption on distinguishing between local (keywords - or more specific document theme) and global (more general) latent topics.
\paragraph{The effects of noises on embedding quality}
We argue that the main reason why the current clustering task is challenging for all observed models, specifically unsupervised ones is mainly due to the \emph{noisy} characteristic of cQA archives. For instance, topic ''Health'' and ``Dining out'' may both contain queries about dietary or source of healthy food. Topic ``Cars'', ``Travel'', ``Local Business'' may all contain queries about car rental and service. We empirically show that in a clean scenario -- where training examples only contain $N$-most influential words w.r.t. topic category (Toy data experiment in fig.\ref{fpr:toy}) unsupervised methods sufficiently perform well. The performance, however, degrades in the occurrence of noises (larger pre-selected feature size). By contrast, \textbf{KeyGen2Vec} can maintain its considerably high performance (fig.\ref{fpr:toy}-\ref{fpr:11t}) regardless the presence of noises. This indicates the exposure of keywords as sub-topical information benefits the model to obtain high quality embeddings. 


\paragraph{Problem reformulation improves the expressiveness of embeddings}
Redefining one-to-many multi-label learning into one-to-one multi-class learning scheme via Softmax normalization, which we argue is analogous to indirectly incorporating label dependency (sec. \ref{sec:dependency}), benefits \textbf{KeyGen2Vec} and Multi-label learning in the current study, resulting in a more accurate embedding (higher $F_1$-score, in table \ref{tab:cluster_toy}-\ref{tab:cluster_11t_qa} and fig.\ref{fpr:toy}-\ref{fpr:11t}).

\section{Conclusion}


We extensively investigate document embedding approaches for topical clustering of cQA archives. We show current limitations of unsupervised embeddings on dealing with noisy articles, indicating the need of incorporating strong assumption either on learning approach or data. Our empirical results highlight the capability of the proposed \textbf{KeyGen2Vec} in preserving topical proximity in latent space via multi-label multi-class learning.


\bibliography{anthology,emnlp2020}
\bibliographystyle{acl_natbib}

\appendix

\section{Document Embedding Methods}
\label{sec:dc_intro}

\subsection{Non-distributed approach}

\paragraph{Frequency-based (TfIdf)} 

\emph{TfIdf} \cite{salton1975vector} -- also referred to as ``bag-of-words'' model, is commonly used as a standard approach to transform text document into numerical representation. A document is represented as vector of semantics, where each dimension reflects a degree of importance (based on relative frequency-based weight) of a particular word $w$ in the corresponding document $d_i$ and across documents in corpus $ d_i \in D$, $w \in \mathcal{V}$ -- where $\mathcal{V}$ denotes vocabulary size of the corpus.

\subsection{Probabilistic approach}

\paragraph{Probabilistic Topic Model}
Latent Dirichlet Allocation (LDA) \cite{blei2010topic,blei2012probabilistic} -- commonly known as Topic Model, is a probabilistic mixture model that views a document as a collection of un-ordered words (``bag-of-words''). The document is represented as a mixture membership of topics - where each dimension of the document vector corresponds to a probabilistic distribution of topic-$c_j$ in the corresponding document, $c_j \in \mathbb{C}$. Each topic is represented as a vector of most probable words defining the topic.

\subsection{Distributed approach}

\paragraph{Mean Embedding} -- a bottom-up approach. A document is viewed as a collection of bag-of-word embeddings. The word-level embedding is learnt through non-linear mapping of neural network architecture \cite{mikolov2013distributed,mikolov2013efficient}. The document representation is computed based on averaging word-level representations occur in the corresponding document. We acquire word embedding $w \in \mathcal{V}$ through both pretrained models, i.e. Word2Vec pretrained on Wikipedia corpus \cite{yamada2020wikipedia2vec}, GloVe vector \cite{pennington2014glove}, and trainable model -- i.e. by training the Word2Vec model \cite{mikolov2013distributed,mikolov2013efficient} and Pointwise Mutual Information (PMI) - based embedding on the current corpus. In table \ref{tab:all_models}, the pretrained models of word embedding are represented by \texttt{\small \{Avg-GloVe100, Avg-w2v100, Avg-GloVe300, Avg-w2v300\}}. The trainable models are represented by \texttt{\small \{Avg-w2v50-tr-sm, Avg-w2v50-tr-lg, Avg-PMI50\}}.

\paragraph{Document Covariance Matrix} -- a bottom-up approach, represented by \texttt{\small \{DC-GloVe100, DC-w2v100, DC-PMI50, DC-w2v50-tr-lg\}} in table \ref{tab:all_models}. A document is represented as the multivariate gaussian embedding (covariance matrix) of its word-level representation 
\cite{nikolentzos-etal-2017-multivariate,torki-2018-document}. To construct a document covariance matrix, we use two types of word-level representations: (1) Pointwise Mutual Information (PMI) \cite{levy2014neural} that measures the association of words $w$ and their context $c$ by calculating the cooccurrence of words and their neighboured words in sentence or document; (2) Word embedding \cite{mikolov2013distributed} that learn distributed representation of words-context words via a neural network architecture - or log linear mapping function (\emph{here}, we use both pretrained and trainable word embedding models).

\paragraph{Paragraph Vector} -- represented by \texttt{\small \{D2V-DBOW100, D2V-PVDM100\}} in table \ref{tab:all_models}. Document and the corresponding words are mapped into a share vector space -- where the objective is to predicting target words, given document and context words in Distributed Memory model \textbf{(PV-DM)}; and to predict context words, given a document in Distributed Bag-of-Words model \textbf{(PV-DBOW)} \cite{le2014distributed}.

\paragraph{Multi-class classifier}
We utilize neural network with dense connections (MLP, denoted as \texttt{\small FC-Mult-Cls} in table \ref{tab:all_models}), representing an \textbf{upper-bound} model in this study. Discrete global class structure in the observed data sets $\mathbb{C}$ is exposed as training objective to learn and condition document features from this MLP-based model. The network architecture is composed of an embedding layer, dropout networks, a pooling layer as a flattening mechanism, and a stack of two fully-connected (FC) networks. The objective of the study, thus is to find the best feature extractor that closer to the quality of features based Multi-class classifier.

\paragraph{Multi-label classifier}
To provide a \emph{fair} comparison with the proposed Seq2Seq framework, we utilize a Multi-label classifier based on dense networks (\texttt{\small FC-Mult-Lbl} in table \ref{tab:all_models}) to model document features condition on multiple dependent labels. Compared to Multi-Class classifier (\texttt{\small FC-Mult-Cls}) that holds independent assumption of $X \mapsto Y$ mapping tasks, Multi-label classifier sees the tasks as mutually inclusive, as such one document can correspond to multiple labels (e.g. tags, keywords). Compared to Seq2Seq that learn to predict a set of sequences based on tree-based mutually dependent structure $Y \in \Sigma^*$ , the Multi-label classifier estimates the probability of multiple classes for one source instance independently (\emph{sigmoid}, instead of normalized \emph{softmax} probability outputs).

\paragraph{Pretrained Model: BERT}
BERT \cite{devlin-etal-2019-bert} is the most recent language representation model surprisingly performed well in diverse language understanding benchmark - indicating the network has a capacity to capture structural information from natural language data \cite{xu-etal-2019-bert, jawahar-etal-2019-bert, reimers-gurevych-2019-sentence}. Unlike its deep architecture counterparts commonly composed of recurrent network to hold the main assumption of sequential data (e.g. Seq2Seq), BERT network is mainly composed of dense connections - referred to as ``self-attention'' network. We utilize a pretrained BERT as static universal sentence encoder -- i.e. to transform documents into vectors unsupervisedly without fine-tuning, assuming the global and sub semantic structure is \emph{unobserved}. We use the latest implementation of BERT for sentence embedding (S-BERT) \cite{reimers-gurevych-2019-sentence}, which has shown a better generalization performance than vanilla BERT \cite{devlin-etal-2019-bert}.

\section{Seq2Seq Networks}

\subsection{Attention Network}

\begin{figure}[!ht]
    \centering
    \includegraphics[scale=.25]{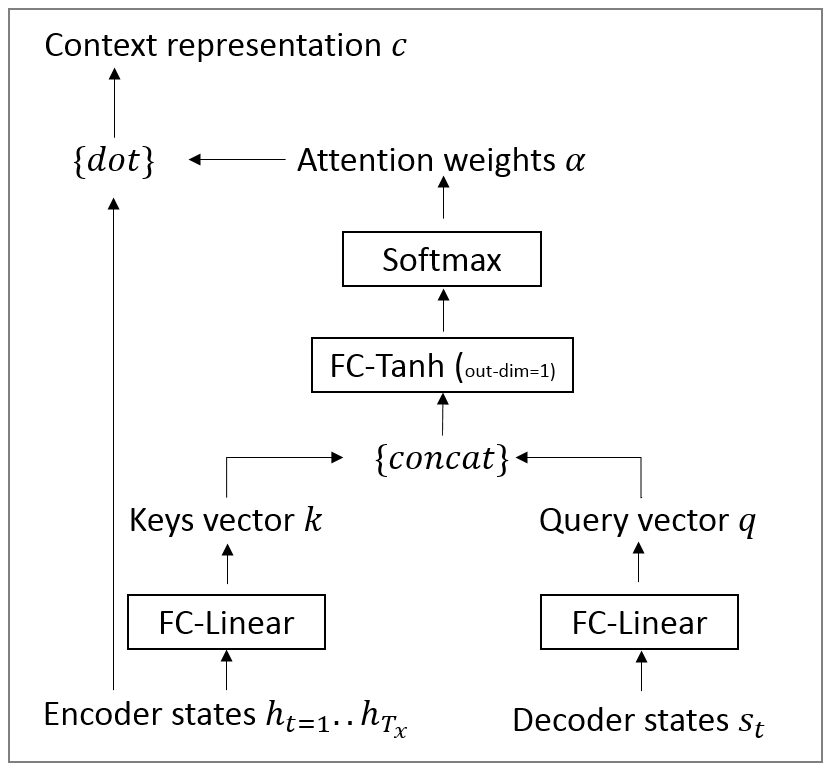}
    \caption{Attention network \cite{Bahdanau2014Neural}.}
    \label{fig:at_bahdanau}
\end{figure}

We use Bahdanau's concat attention scoring function \cite{Bahdanau2014Neural}, illustrated in fig. \ref{fig:at_bahdanau}, to calculate attention weights $\alpha \in R^{T_x}$ of encoder state representation conditioned by decoder output state representation.

\subsection{Teacher forcing}
\label{sec:teacher-forcing}
A common strategy to train a recurrent-based Seq2Seq model in a generation task is incorporating teacher forcing, i.e. by exposing the actual or expected output $Y_{t<T_y}$ at the current decoding time step $t<T_y$, rather than the output generated by the network ($s_{t<T_y}$). The drawback, however, during evaluation stage the network only relies on its own prediction from previous time steps, resulting a performance degradation referred as ``training-evaluation loss mismatch'' in \cite{wiseman2016sequence}.
In this study, we employ curriculum learning \cite{Bengio2015Scheduled}, i.e. approach to sampling whether to use teacher forcing or not during training stage in the current sequence prediction problem. Here, the probability of incorporating teacher forcing (ratio of teacher forcing) is calculated based on inverse sigmoid function of scheduled sampling (after $xx-$ batch examples seen during training).

\section{Evaluation Metrics}

\paragraph{Clustering Evaluation}
A normalized mutual information (NMI) is used to measure whether the clustering method can recreate the true or exact structure of the observed data. We use the following metrics to evaluate the structure representation in a clustering task, in addition to $F_1$-score and purity measure.

\begin{equation}
    NMI(\Omega, \mathbb{C}) = \frac{I(\Omega, \mathbb{C})}{0.5(H(\Omega)+H(\mathbb{C}))}
\end{equation}

Where $I(\Omega,\mathbb{C})$ measures the mutual information between the formed cluster membership and exact class. And, $H(\Omega), H(\mathbb{C})$ denotes entropy of the formed structure and exact class respectively. 

\begin{equation}
    I(\Omega, \mathbb{C}) = \sum_k \sum_j P(\omega_k \cap c_j) \log \frac{P(\omega_k \cap c_j)}{P(\omega_k) P(c_j)}
\end{equation}

\begin{equation}
    H(\mathbb{C}) = -\sum_k P(\omega_k) \log P(\omega_k)
\end{equation}

$P(\omega_k), P(c_j), P(\omega_k \cap c_j)$ are the probability of document feature being in cluster $\omega_k$, class $c_j$, and both memberships.

\paragraph{Purity}

Each cluster is assigned to the class which has the most frequent members in the cluster, as such the purity of clusters is computed by $\texttt{purity}(\Omega, \mathbb{C})=\frac{1}{N}\sum_k \texttt{max}_j|\omega_k \cap c_j|$. $\Omega = \{ \omega_1, \omega_2, \ldots, \omega_k \}$ denotes a set of $k$ clusters, while $\mathbb{C} = \{ c_1, c_2, \ldots, c_j \}$ is a set of $j$ actual classes. A perfect clustering has a purity of $1$. \textbf{Note}: the purity metric disregards the \emph{uniqueness} of the cluster since it is computed based only on the majority class and number of members of the majority class.

\paragraph{Normalized Mutual Information (NMI)}
NMI or a Mutual Information-based metric measures the amount of information needed to predict the actual class of a cluster, given a knowledge about documents in that cluster. 

\vspace{-.5em}
\begin{align*}
    NMI(\Omega, \mathbb{C}) = \frac{I(\Omega, \mathbb{C})}{ [H(\Omega) + H(\mathbb{C})] /2} \\
    I(\Omega, \mathbb{C}) = \sum_k \sum_j P(\omega_k \cap c_j) log \frac{P(\omega_k \cap c_j)}{P(\omega_k) P (c_j)}
\end{align*}

\noindent where $I$ is mutual information between the predicted clusters $\Omega$ and the actual classes $\mathbb{C}$, which is measured based on the overlapping document membership between clusters and actual classes. $H$ denotes an entropy measure $H(P)=-\sum_{x \in X} P(x) log_2 P(x)$.


\paragraph{F1-Score}
\label{sec:f1-metric}
We measure F1-score (harmonic mean of Precision and Recall) of document clustering based on the notion of how a pair of documents (points in latent space) is assigned into clusters \cite{manning2008introduction}. \textbf{True Positive (TP)} assignment implies that two \emph{similar} documents are assigned to the same cluster. \textbf{True Negative (TN)} assignment refers to the assignment of two \emph{dissimilar} documents to different clusters. \textbf{False Positive (FP)} assigns two dissimilar documents into the same cluster, while \textbf{False Negative (FN)} assigns two similar documents into different clusters. Precision (P) and Recall (R) are then computed based on $\small P = TP/(TP+FP)$ and $\small R = TP/(TP+FN)$; while $\small F_1$-score is a harmonic mean of both metrics $\small F_1 = (2PR)/(P+R)$.


\section{Data}

\begin{table}[!ht]
    \centering
    \resizebox{.5\textwidth}{!}{
    \begin{tabular}{p{4cm}|p{2.5cm}|p{1.5cm}}
    \hline
    Sentence & Keywords  & Topic  \\
    & (SUB) & (GLO) \\
    \hline
    ``what was so important about the battle of quebec'' & battle of quebec; american revolutionary war & history \\
    \hline
    ``who were the commanders that died in the battle of quebec in 1759'' 
     & battle of quebec; american revolutionary war & history \\
     \hline
     ``the sporozoan plasmodium carried from host to host by mosquitoes causes what serious infection'' & malaria; plasmodium parasite & virus and diseases \\
     \hline
     ``plasmodium is a malaria causing sporozoan which is transmited by mosquito'' & malaria; plasmodium parasite & virus and diseases \\
    \hline
    \end{tabular}
    }
    \caption{Sentence examples in Toy Data set}
    \label{tab:toy_examples}
\end{table}

\begin{table}[!ht]
    \centering
    \resizebox{.5\textwidth}{!}{
    \begin{tabular}{p{8cm}}
    \hline
    ``diabetes mellitus is medical disorder characterized by varying or persistent hyperglycemia elevated blood sugar levels, especially after eating. all types of diabetes mellitus share similar symptoms and complications at advanced stages. hyperglycemia itself can lead to dehydration and ketoacidosis. longer term complications include cardiovascular disease doubled risk, chronic renal failure it is the main cause for dialysis, retinal damage which can lead to blindness, nerve damage which can lead to erectile dysfunction impotence, gangrene with risk of amputation of toes, feet, and even legs''  \\
    \hline
    \textbf{keywords:} diabetes; diseases and conditions \\
    \textbf{topic:} health \\
    \hline
    \end{tabular}
    }
    \caption{Sentence examples in Yahoo! Answer Q\&A}
    \label{tab:qa_yahoo_examples}
\end{table}

\begin{figure*}[!t]
     \centering
     \begin{subfigure}[ht]{0.2\textwidth}
         \centering
         \includegraphics[width=1in,height=1in]{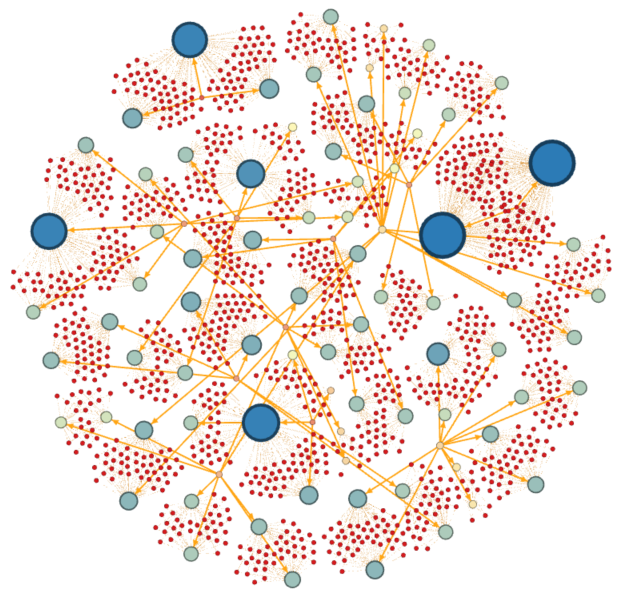}
	\caption{}
	\label{net-a}
     \end{subfigure}
     \begin{subfigure}[ht]{0.2\textwidth}
         \centering
         \includegraphics[width=1in,height=1in]{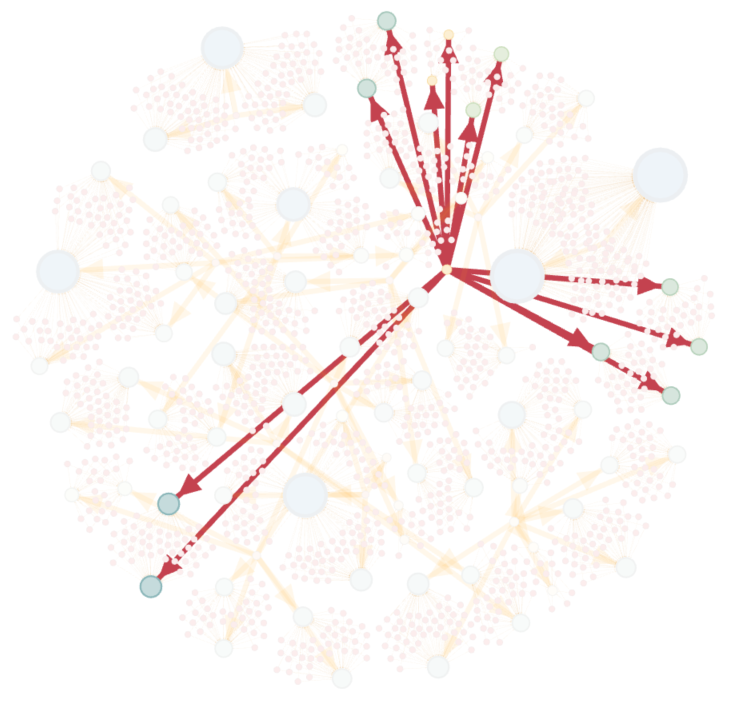}
	\caption{}
	\label{net-b}
	\end{subfigure}
	\begin{subfigure}[ht]{0.2\textwidth}
         \centering
         \includegraphics[width=1in,height=1in]{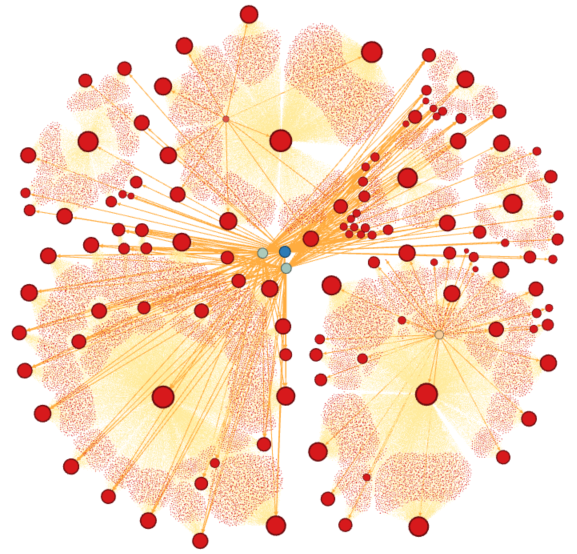}
	\caption{}
	\label{net-c}
     \end{subfigure}  
	\begin{subfigure}[ht]{0.2\textwidth}
         \centering
         \includegraphics[width=1in,height=1in]{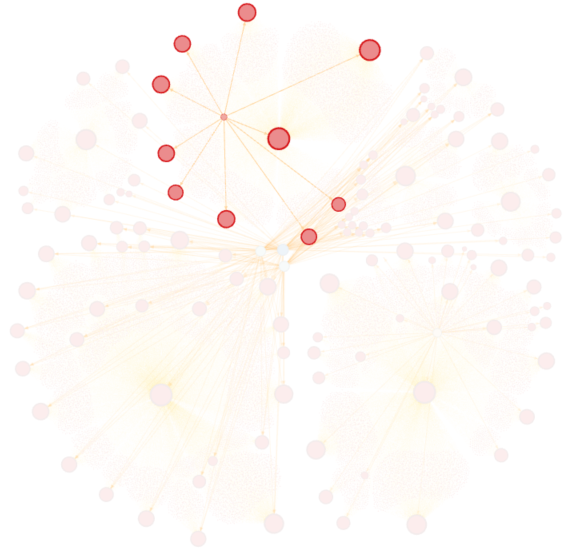}
	\caption{}
	\label{net-d}
     \end{subfigure}
	\begin{subfigure}[ht]{0.2\textwidth}
         \centering
         \includegraphics[width=1in,height=1in]{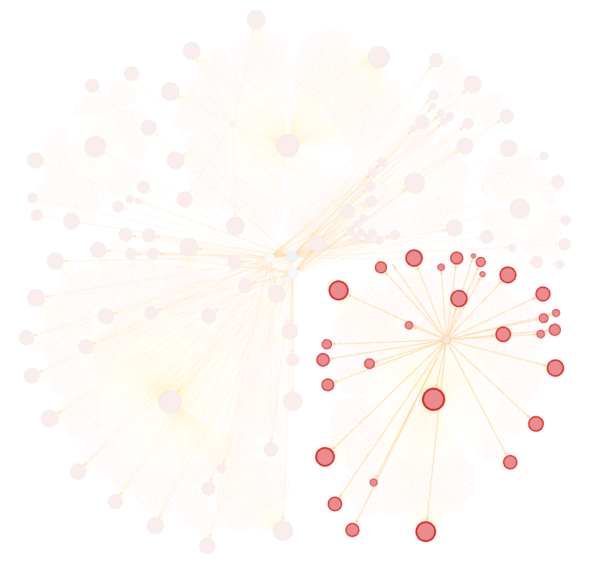}
	\caption{}
	\label{net-e}
       \end{subfigure}
	 \begin{subfigure}[ht]{0.2\textwidth}
         \centering
         \includegraphics[width=1in,height=1in]{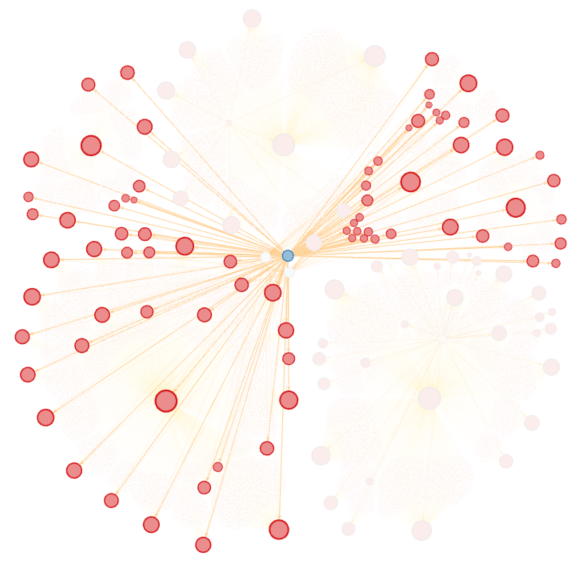}
	\caption{}
	\label{net-f}
       \end{subfigure}
	\begin{subfigure}[ht]{0.2\textwidth}
         \centering
         \includegraphics[width=1in,height=1in]{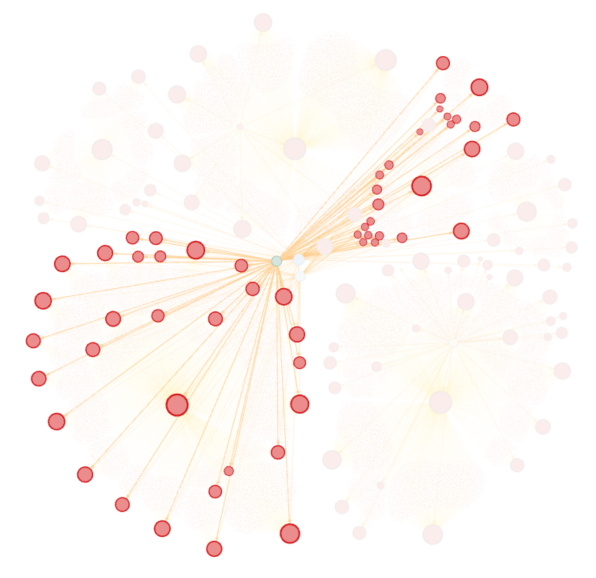}
	\caption{}
	\label{net-g}
        \end{subfigure}
	\begin{subfigure}[ht]{0.2\textwidth}
         \centering
         \includegraphics[width=1in,height=1in]{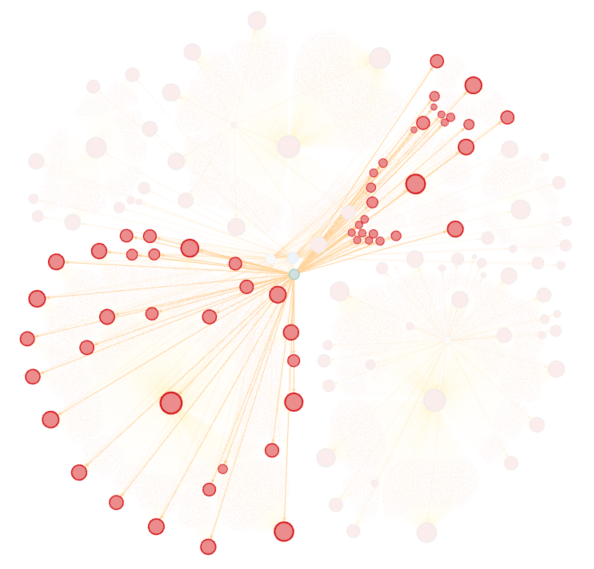}
	\caption{}
	\label{net-h}
      \end{subfigure}
        \caption{Corpus visualization as graph network (nodes represent documents, keywords, and topic categorization); $(a-b)$ Toy data; $(c-h)$ 5-T Yahoo! cQA data.}
        \label{fig:net_struct}
\end{figure*}
\section{Corpus Visualization}

Figure~\ref{fig:net_struct} shows a corpus visualization as graph network where nodes represent document sources that are connected by keywords as sub-structure and topic labels as global structure.

\section{More Results}

\subsection{KeyGen2Vec as Embedding Models}
\textit{``Under which condition KeyGen2Vec is better than the other embedding models?''}

\paragraph{On Scalability Aspect of Model}
Compared to unsupervised methods and pretrained models, KeyGen2Vec has shown a consistent good performance on capturing semantic structure in data, outperforming the other models, regardless the number of training examples. \emph{Note}: This specifically holds on scale-free network (Figure~\ref{fig:net_struct} and Figure~\ref{fig:net_define}), i.e. the growth of the network is independent with the underlying structure of the network. While word-level embedding and Seq2Seq for autoencoding suffer on small set of training examples (Toy data, where the evaluation result is shown in Table~\ref{tab:cluster_toy}), Seq2Seq for keyword generation (STS-KG) shows an ability to learn useful features even in small data set, indicating the model can be utilized as feature extractor for both small and large scale data.    

\begin{figure}[!ht]
\centering 
\begin{subfigure}[ht]{0.2\textwidth}
\centering
\includegraphics[width=\linewidth]{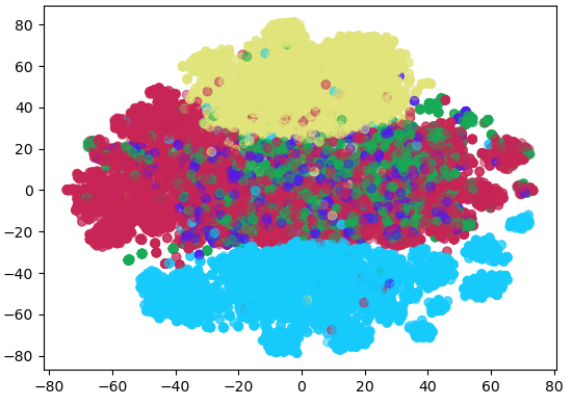}
\caption{}
\label{loss:a}
 \end{subfigure}   
\begin{subfigure}[ht]{0.2\textwidth}
\centering
\includegraphics[width=\linewidth]{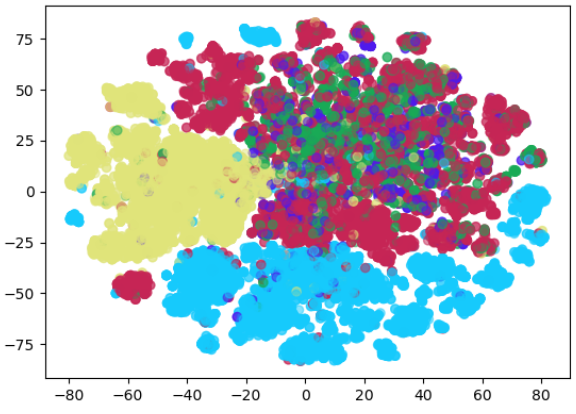}
\caption{}
\label{loss:b}
 \end{subfigure}   
\caption{Loss in approximating global structure is mainly due to overlapping sub-structures; (a) STS-KG; (b) Multi-Label Classifier}
\label{fig:loss}
\end{figure}

\paragraph{On Semantic Structure Inferred in Data}
While sub-structure can promote the learning of \emph{latent} global structure inferred in data, shown in our empirical results, an overlapping sub semantic structures exemplified by the two real world data sets in the current study, as shown in Figure~\ref{net-c}-\ref{net-h}, may potentially introduce noises in the learning. This type of loss is also shown on the extracted features from both models that learn representation from sub semantic structures: Seq2Seq for keyword generation STS-KG (Figure~\ref{loss:a}) and multi-label classifier (Figure~\ref{loss:b}). The overlapping points in different colors (Figure~\ref{fig:loss}) represent documents in three category topic labels: \texttt{\footnotesize 'dining out', 'travel', 'local business'} that shares common set of keywords (\texttt{\footnotesize ``London'', ``UK''}).

\begin{figure*}[!ht]
     \centering
	\begin{subfigure}[ht]{0.25\textwidth}
         \centering
         \includegraphics[width=\linewidth]{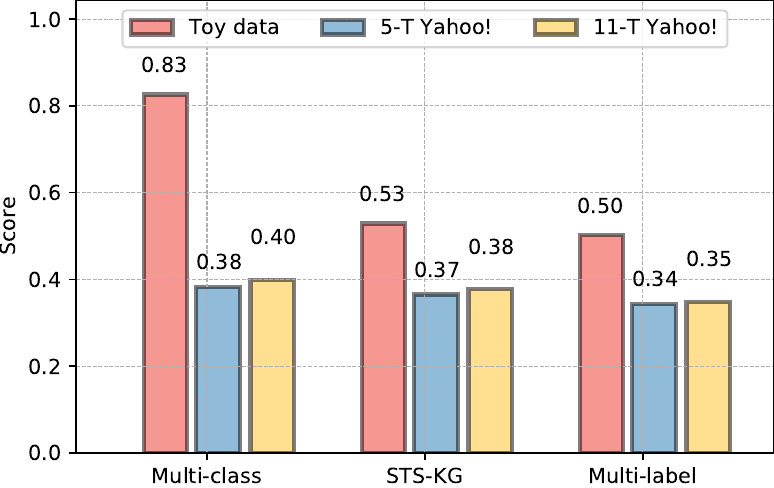}
	\caption{}
	\label{silhouette:a}
        \end{subfigure}   
	\begin{subfigure}[ht]{0.2\textwidth}
         \centering
         \includegraphics[width=\linewidth]{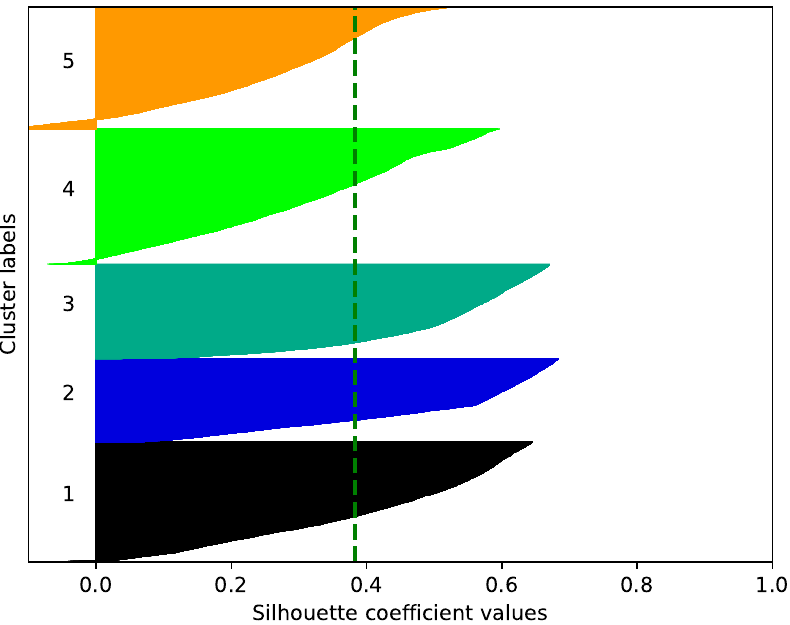}
	\caption{}
	\label{silhouette:b}
        \end{subfigure}   
	\begin{subfigure}[ht]{0.2\textwidth}
         \centering
         \includegraphics[width=\linewidth]{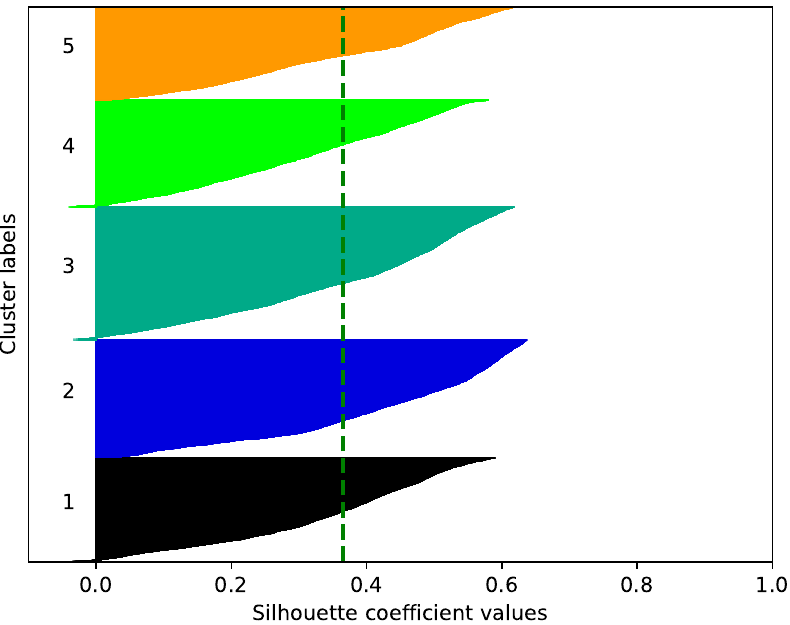}
	\caption{}
	\label{silhouette:c}
        \end{subfigure}   
	\begin{subfigure}[ht]{0.2\textwidth}
         \centering
         \includegraphics[width=\linewidth]{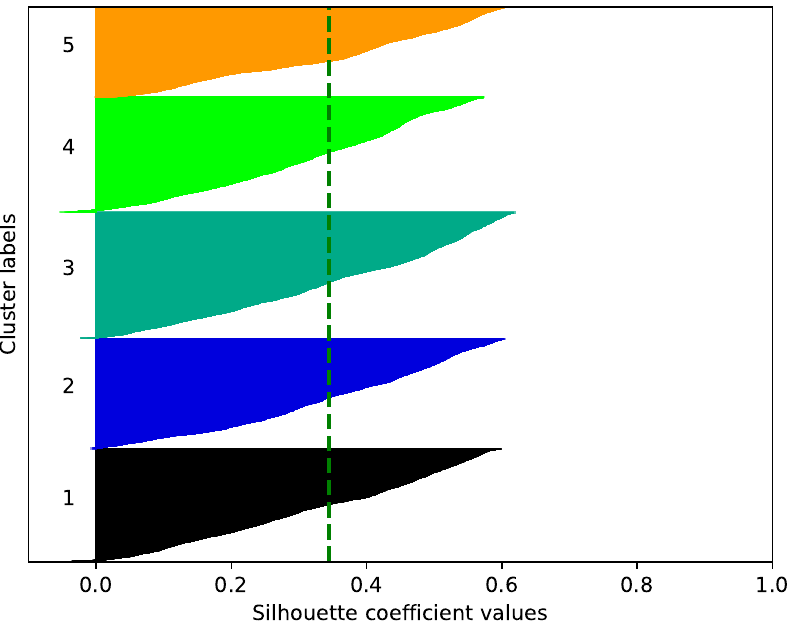}
	\caption{}
	\label{silhouette:d}
        \end{subfigure} 
        \caption{(a) Average silhouette score on three data sets; (b) Multi-Class Classifier on 5-T Yahoo!; (c) STS-KG on 5-T Yahoo!; (d) Multi-Label Classifier on 5-T Yahoo!;}
       \label{fig:silhouette_sc}
\end{figure*}

\paragraph{KeyGen2Vec vs. Seq2Seq Autoencoding}

The clustering evaluation results presented in table \ref{tab:cluster_toy}-\ref{tab:cluster_11t_qa} notably shows that Seq2Seq autoencoder (\textbf{STS-AE}) cannot adequately capture the implicit global semantic structure of data. Similar to the small version of word embedding-based model in the current work (\texttt{\small Avg-w2v-tr-sm} table \ref{tab:all_models}), we argue that an autoencoder heavily relies on the ``goodness'' in the data, indicating the model may be useful if meaningful sentences that promote a content-based semantic structure learning are available abundantly as training examples.  If such case is not available, the performance of feature extractor may be improved by inferring bias to the model architecture or training objective. For instance, instead of an autoencoding task, the network can be trained to predict the semantic structure of the document source, exemplified by Seq2Seq framework for keyword generation in the current study.

\paragraph{KeyGen2Vec vs. Multi-Label Classifier} \hfill

Figure \ref{fig:silhouette_sc} shows the comparison of cluster separability of the three model based on silhouette measure: (1) Multi-class classifier as an \textbf{upper bound} model; (2) Seq2Seq for keyword generator; and (3) Multi-label classifier. The vertical dashed lines in fig \ref{silhouette:b}-\ref{silhouette:d} represents the average score of all points in the resulting clusters, while the size of the bar plot represents the size of cluster. Silhouette score was calculated based on the average distances for all points in the same cluster and the average distances for points in the closest cluster. The score values range from $[-1,1]$ -- where score$=-1$ indicates sample is assigned to the wrong cluster, average score of $0$ indicates that the inter-cluster distance is small, and average score of $1$ indicates the inter-cluster distance is large enough to form separable clusters. Intuitively, the silhouette score is expected to close to $1$ for a good quality of clusters. A good quality of clusters infers a good quality of extracted features. 

Overall, without having the exact classes to evaluate the quality of document clustering, Both features from STS-KG and Multi-Label classifier can approximate the cluster separability of an upper bound model, indicating both models has a capacity to extract ``good'' features according to the current definition of global semantic structure in the current study. Although the clustered features based on Seq2Seq (STS-KG) has a slightly higher score than multi -label classifier clustered features, the actual quality of extracted features by Seq2Seq outperformed multi-label classifier features when exact classes are projected on the resulting clusters.

\paragraph{KeyGen2Vec vs. LDA Topic Model}
\label{sec:lda}

While KeyGen2Vec has outperformed all models in the current study, this performance comes at cost of providing sub semantic structure information. As a comparison, we discuss the performance trade-offs of LDA Topic Model as an alternative model for unsupervised approach, in addition to the aforementioned unsupervised and pretrained models (sec. \ref{sec:unsup}).

\begin{figure}[!ht]
\centering
         \includegraphics[width=\linewidth]{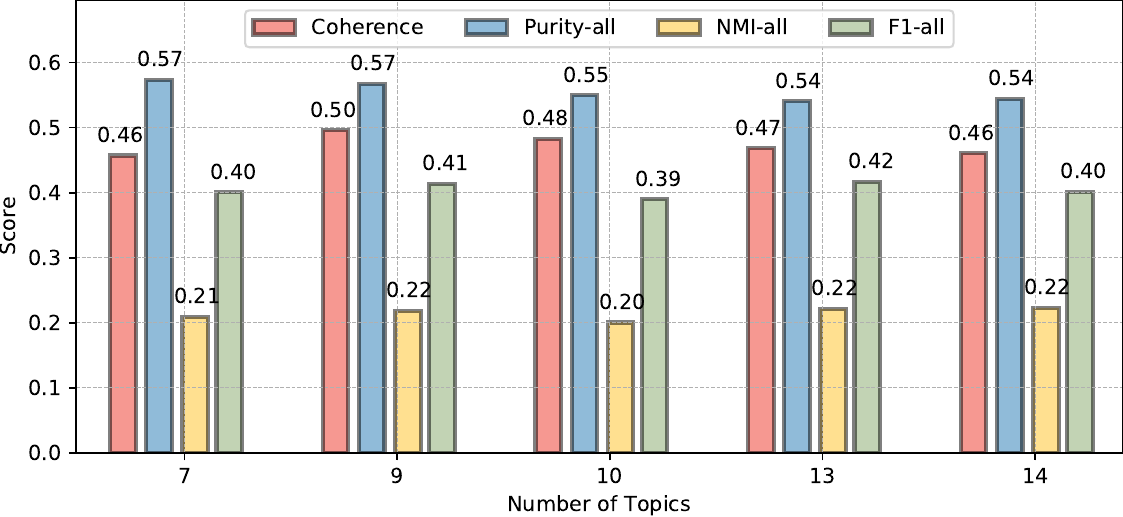}
	\caption{Deciding hypothesis space (number of topics)}
	\label{fig:lda_ntopics}
\end{figure}

\begin{figure}[!ht]
\centering
         \includegraphics[width=.8\linewidth]{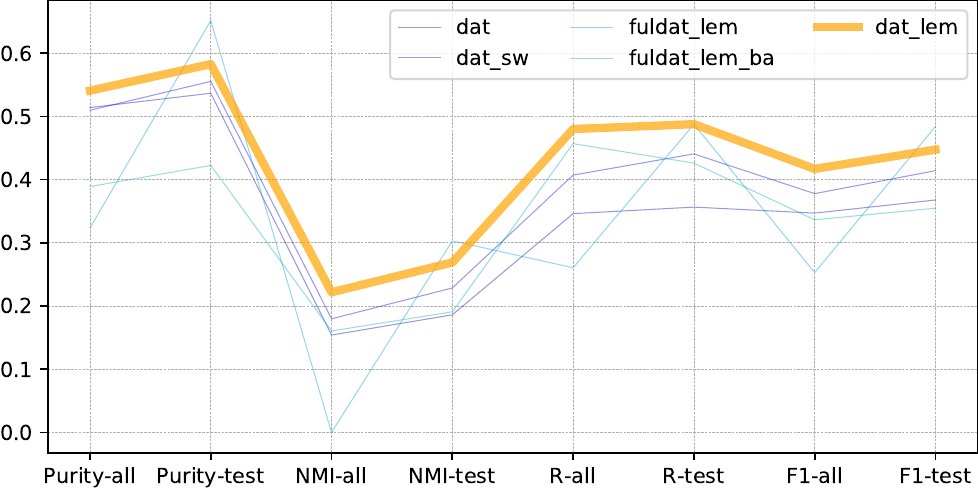}
	\caption{LDA Performance on different preprocessing steps}
	\label{fig:lda_vs_dat}
\end{figure}

While LDA topic model in the current study (results in table \ref{tab:cluster_toy}-\ref{tab:cluster_11t_qa}) does not show an impressive performance w.r.t. the semantic quality of the learnt clusters, the model promotes a  consistent quality as document feature extractor among three data sets, similar to a trainable word embedding-based model (\texttt{\small Avg-w2v-tr}). Nevertheless, deciding a hypothesis space of topic model is non-trivial. Too small $K$ number of topics results in a very broaden topics (words corresponds to the topic semantic definition is too general). While, too large $K$ results in a repetitive topics -- i.e. different topics contain an overlapping set of words. While the hypothesis space of topic model can be evaluated by coherence measure during training, the measure does not necesarily correlate to the actual quality of learnt features or an interpretability aspect of the resulting features in topic space , illustrated in fig. \ref{fig:lda_ntopics}. We refer the ``interpretability'' here as a degree to which the two different clusters are adequately far or \emph{separable}.

The performance of topic model in capturing document semantic structure also highly depends on heavy preprocessing steps. Figure \ref{fig:lda_vs_dat} shows how the quality of extracted features varies, depending on (a) whether the stopword and noisy words have been removed (e.g. removing question words and non meaningful abbreviation or short character in \texttt{\small dat\_sw}); (b) whether data is lemmatized (\texttt{\small dat\_lem}); (c) document length (\texttt{\small fuldat\_lem}); or (d) a balance distribution of topic categorization labels (\texttt{\small fuldat\_lem\_ba}). If such knowledge (e.g. POS tag linguistic structure in lemmatization step) is not available in particular data or language, the performance can be expected to degrade.

\end{document}